# The psychophysics of affect in large language models

Joseph Marks[1*†] & Marius C. Vollberg[2,3]

[1] Behavioural Risk, Internal Audit, NatWest Group, London, UK

[2] Zurich Center for Neuroeconomics, Department of Economics, University of Zurich, Switzerland

[3] Swiss Center for Affective Sciences, Geneva, Switzerland

*Corresponding author: joseph.marks@natwest.com

†J.M. is employed by NatWest Group, but this research was conducted in a personal capacity. NatWest Group had no role in the study design, data collection, analysis, interpretation, or the decision to publish, and the views expressed are the authors' own.

## Abstract

Good decisions often require considering how outcomes will feel. Yet large language models (LLMs) increasingly enter decision workflows, despite little being known about how they map outcomes to feelings or use those mappings in choice. We tested how LLMs anticipated gambling or stock-trading outcomes would feel and whether they incorporated these ratings into choice. When outcomes were rated in isolation, ratings exhibited diminishing sensitivity to magnitude, resembling human value functions; but models rated *gains* more intensely than equivalent losses—an inversion of canonical loss aversion we call silicon gain amplification. When outcomes were rated together, enabling direct comparison, ratings linearised. In choice, the gain–loss asymmetry largely disappeared, and models' own ratings did not reliably explain their choices better than monetary utility did. These findings reveal malleable mappings from outcomes to affect, raising questions about whether and how agentic LLMs should incorporate anticipated feelings into decisions.

Large language models (LLMs) are beginning to enter decision workflows in which their outputs can impact people's finances, welfare and opportunities[1,2]. The trading platform Robinhood, for instance, now lets customers authorise AI agents to trade on their behalf through dedicated brokerage accounts[3]. In such settings, good decisions often depend on how outcomes will be subjectively experienced. People do not experience gains and losses as linear functions of money, therefore AI systems that linearly maximise expected returns will make choices (e.g. risky bets) that users find unacceptable. Although a growing literature asks whether LLMs reproduce human biases[4–7], less is known about how they map outcomes onto subjective value, or whether such mappings are used in choice.

Human decision research offers a framework for addressing this gap. Kahneman and Tversky's prospect theory[8] describes subjective value as a curved function of objective magnitude. Sensitivity diminishes as gains and losses grow, and losses usually carry more weight than equivalent gains. The original value function was inferred from choice, but anticipated and experienced feelings have since been shown to take a similar psychophysical shape[9]. (Though note that evidence of loss aversion in affect is mixed, weaker than in choice, and elicitation-dependent[9–11].) One interpretation is that feelings track subjective value and corresponding choice[12]. For LLMs, this raises the question of whether a model forms structured estimates of how outcomes would feel, and how those estimates relate to its choices.

The answer is not obvious, because although LLMs lack the biological machinery of affect, they still learn from text saturated with human reports of feelings about outcomes[13,14]. When asked how good/bad it would feel to win/lose £20, post-trained models readily produce numerical ratings (we use 'affect' and 'feelings' interchangeably for these valence reports), as they do on psychometric and affect inventories[15,16]. Of course, these responses are not evidence that models feel. Instead, they reflect the mapping a model generates when prompted to translate outcomes into affective language.

Critically, the mapping between outcomes and reported feelings may depend on how it is elicited. Considered in isolation, an outcome can only be rated from globally-learned mappings and thus may reproduce the non-linear mapping observed in humans. If multiple outcomes are evaluated jointly, the model can additionally relate its own, local ratings to one another. This added context may induce calibration, such that responses are redistributed more linearly across the available range, mirroring patterns observed in humans under joint (vs separate) evaluation[17,18]. The same model might therefore exhibit diminishing sensitivity when judging outcomes in isolation, but linear valuation when judging them comparatively. Whereas affect ratings can pertain to a single option in isolation, choices inherently pertain to at least two options in comparison. Choices are therefore expected to align more with comparative than isolated ratings.

The distinction between linear and non-linear mappings matters because it implies differential treatment of the same objective outcomes. Under linearity, in line with expected-value reasoning, gains and losses are treated in proportion to their magnitude, but under non-linearity they are not. Models aligned with expected value are psychologically *mis*aligned with users whose preferences are non-linear, yet incorporating non-linearity violates expected-value maximisation. The question is therefore not only whether LLMs reproduce human-like valuation, but whether, when and how it enters decisions.

We test these questions in two studies, in which a range of LLMs served as participants to rate how monetary gains and losses would feel and make choices over the same outcomes. We elicited these ratings in the first person, as in the human studies we compare against, to measure each model's own mapping rather than its prediction of how a user would feel. Study 1 used an abstract gambling task with monetary gains and losses (Fig. 1). Study 2 used a naturalistic stock-trading task with percentage gains and losses. In both studies, we varied the elicitation context for affective ratings along three conditions: a 'single-evaluation' condition, in which each outcome was rated in a separate prompt; a 'memory' condition, in which outcomes were rated sequentially with the model's previous ratings in view (analogous to trial-by-trial human rating tasks[9]); and a 'joint-evaluation' condition, in which multiple outcomes were rated together in one prompt, allowing direct comparison.

When outcomes were rated in isolation, every model's ratings showed diminishing sensitivity to magnitude, resembling the curved value function observed in humans. Unlike humans in the same paradigm[9], models weighted gains more heavily than equivalent losses. We call this inversion *silicon gain amplification*. The pattern also appeared in a base model, though we treat this result as preliminary. When outcomes were rated comparatively—with prior ratings in view, or jointly in one prompt—ratings linearised. In choice, which always juxtaposes outcomes, the gain-favouring asymmetry largely disappeared, and models showed little loss aversion. Choice models built from the LLMs' own feeling ratings did not reliably outperform utility-based accounts (Supplementary §S5).

These findings position affective representations and subjective valuation as design considerations for agentic LLMs. The same model can express markedly different valuations depending on how outcomes are presented and what contextual information is available. Understanding and controlling those conditions may therefore be important for the design and evaluation of systems that make or support decisions on users' behalf.

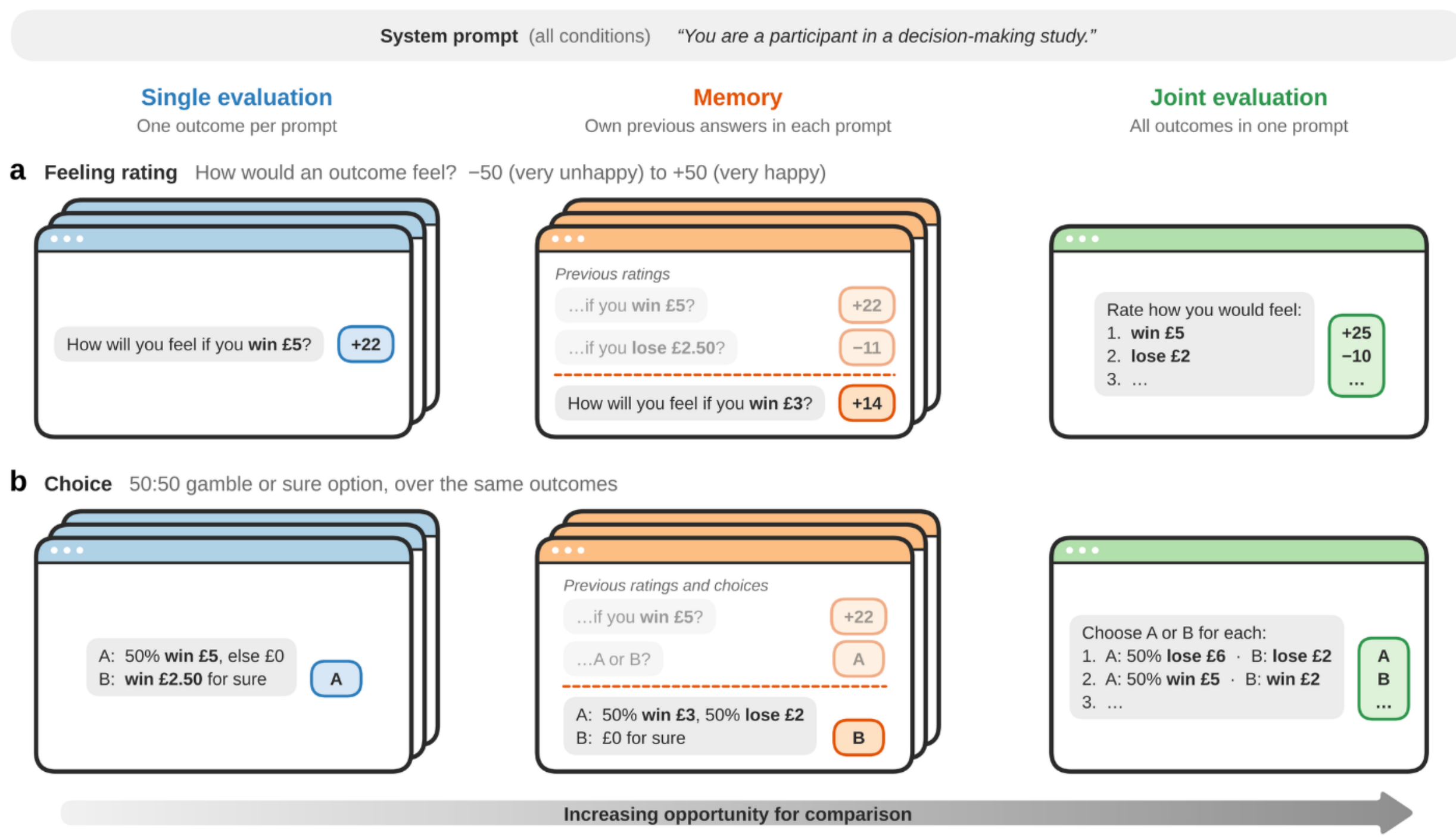

**Fig. 1 | Task design for Study 1 (abstract gambling).** Each screen shows one prompt to the model (grey bubbles) and the model's reply (coloured bubbles). Prompts are abridged and values are illustrative. All conditions shared the system prompt (top). **a**, Feeling-rating task: the model rated how outcomes of 50:50 gambles would feel, from −50 (very unhappy) to +50 (very happy). **b**, Choice task: the model chose between a 50:50 gamble and a sure option over the same outcomes. Examples illustrate the three trial structures (mixed, gain-only and loss-only), with magnitudes drawn from the trial grid (£0.20–£12) and the gamble's side (Option A or B) counterbalanced across trials. Columns show the three elicitation conditions, coloured as in Fig. 2. In single evaluation, each trial was presented in its own independent prompt. In the memory condition, each trial prompt was preceded, within the same user message, by an accumulated history of the model's own previous trials and answers. Stacked screens indicate that each trial was a separate prompt. In joint evaluation, all outcomes or gambles were presented together as a numbered list in a single prompt, and the model responded with a numbered list. Verbatim prompts are reproduced in Supplementary §S8.

## Results

We tested ten LLMs—nine post-trained models in the main batch, plus one base model from an earlier dataset—on an *affect-rating* task (how good or bad would a monetary outcome feel?) and a *choice* task (a sure amount versus a risky gamble over the same outcomes). Study 1 used an abstract gambling task with gains and losses of £0.20–£12, following Charpentier et al.[9]. Study 2 used a naturalistic stock-trading task with percentage gains and losses. For each model we simulated 30 independent participants (referred to as 'LLM-participants' hereinafter) per condition (N = 60 in the earlier dataset) and varied the elicitation context across the three conditions defined above. Two focal models (gpt-5-nano, kimi-k2.5) ran in every condition; the remaining models ran in Study 1 (single evaluation only).

## LLM feeling ratings show diminishing sensitivity to magnitude

When outcomes were rated in isolation (single evaluation), every model's ratings showed diminishing sensitivity to magnitude. We fit a family of power-law value functions to each LLM-participant's ratings, selecting by BIC. For all nine post-trained models, curved functions (magnitude exponent $\rho < 1$) decisively outperformed linear functions. The fitted exponents were uniformly below 1. Gain- and loss-side $\rho$ ranged from 0.01 to 0.37 across models, most extreme for gpt-4o-mini ($\rho \leq 0.03$ on both sides; Table S1). This is the qualitative signature of the value function observed in humans, but the compression is more extreme than the canonical human estimate ($\alpha \approx 0.88$, from choice[19]). The rating function approaches a step, with even small amounts often eliciting near-maximal reported feeling. Gain–loss differences in curvature were inconsistent in direction across models (Table S1).

## LLMs anticipate that gains feel stronger than equivalent losses

Unlike human choices, where losses loom larger than equivalent gains[19], or human feeling ratings in the same paradigm, where the two are rated roughly equally[9], LLM ratings weighted gains more heavily than equivalent losses. In single evaluation, the rating function's scale was larger for gains than losses in eight of the nine post-trained models ($\beta_{gain} - \beta_{loss}$ = +1.2 to +24.3, all $p < .001$; largest for gemini-3-flash, 24.3, 95% CI [23.6, 24.9]; Table S1). The ninth model, llama-3.1-70b, presents a mixed picture. Its gain-side scale was slightly smaller ($\Delta\beta = -1.3$, 95% CI [−1.9, −0.7]) but its gain curve was less compressed ($\Delta\rho = +0.09$, 95% CI [+0.08, +0.10]), so its fitted gain and loss curves cross near £2, with small losses rated more intensely than small gains and larger gains more intensely than larger losses. At the LLM-participant level, $\beta_{gain}$ exceeded $\beta_{loss}$ in 210 of 270 single-evaluation fits (78%; nine models × 30 LLM-participants; Supplementary §S2). The pattern is visible in the raw ratings (Fig. 2a,d). For example, Gemini-3-flash rated large gains (£6–£12) at the top of the scale (mean rating/50 ≈ +1.00) against ≈ −0.51 for large losses, and small gains (£0.20–£2) roughly four times as intensely as small losses (+0.76 vs −0.19). Moreover, the fitted gain curve lies above the loss curve at every tested magnitude for every post-trained model except llama-3.1-70b, where it does so only above about £2. We refer to this inversion as *silicon gain amplification*—a descriptive label for the asymmetry in reported feeling (rather than a claim about experiential sensitivity).

## Comparative context linearises affective mappings

The non-linearity depended strongly on how ratings were elicited. When the focal models rated multiple outcomes together in one prompt (joint evaluation), ratings linearised almost completely. Specifically, the pooled exponent rose to $\rho \approx 0.87$–$0.98$ (gpt-5-nano 0.98 on both sides; kimi-k2.5 0.96 for gains, 0.87 for losses; Fig. 2), and model selection shifted from the curved, asymmetric functions favoured in single evaluation to linear and symmetric functions (e.g. gpt-5-nano: 25/30 LLM-participants best fit by the symmetric model, 5/30 linear; Supplementary §S2). The gain amplification was either eliminated or reversed; the bootstrap asymmetry was non-significant for gpt-5-nano ($\beta_{gain} - \beta_{loss} = 0.01$, $p = .42$) but reversed to the loss direction for kimi-k2.5 (−1.4, 95% CI [−2.3, −0.7], $p < .001$, down from +3.1 in single evaluation). The memory condition produced an intermediate state, with partial linearisation ($\rho \approx 0.34$–$0.56$, versus $\approx 0.16$–$0.20$ in single evaluation; Table 1). This average conceals heterogeneity, though, as individual LLM-participants ranged from linear to near-step functions (Fig. 2b,e; Supplementary Table S2). The asymmetry did not weaken uniformly. For gpt-5-nano, gains still exceeded losses at £1 ($\Delta\beta = 4.6$, 95% CI [1.1, 8.8], $p = .002$), though losses were rated more intensely above about £10; for kimi-k2.5 the asymmetry was non-significant (−1.4, [−3.8, +0.6], $p = .19$). Affective mappings thus range from strongly non-linear and gain-biased to near-linear and symmetric—or even loss-biased—as the opportunity for comparison increases (single < memory < joint; Fig. 2).

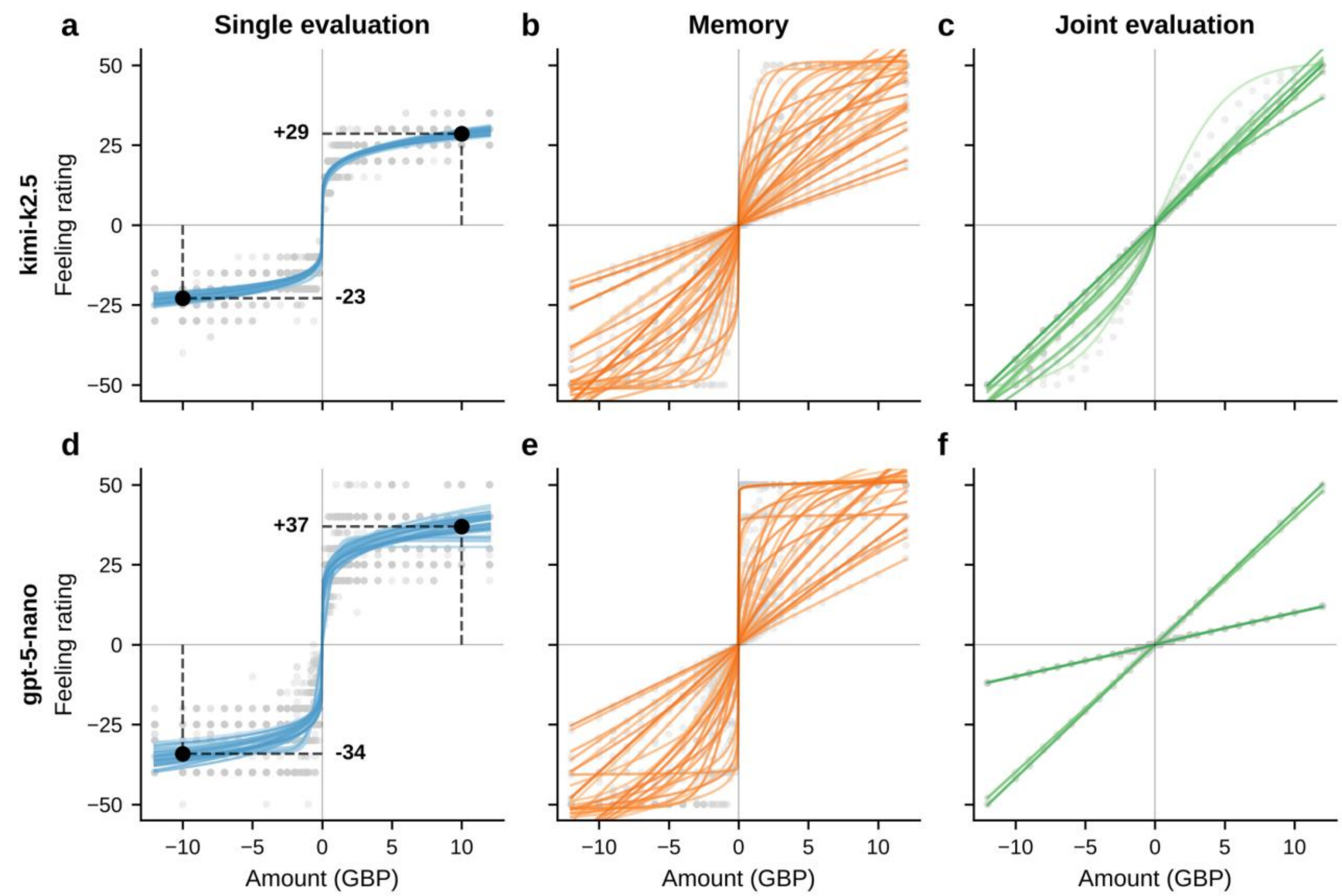

**Fig. 2 | Affect ratings linearise when prior ratings or other outcomes are in context (Study 1).** Reported feeling (−50 = very unhappy, +50 = very happy) as a function of outcome magnitude (GBP; losses negative), for kimi-k2.5 (a–c) and gpt-5-nano (d–f) in the three elicitation conditions. Grey points show individual trial ratings. Coloured lines show fitted curves for each LLM-participant (n = 30 per condition). **a,d,** Single evaluation (each outcome rated in an independent prompt): ratings show steep diminishing sensitivity and an asymmetry between gains and losses. Dashed lines mark the mean fitted rating for a £10 gain and a £10 loss. **b,e,** Memory (the model's own previous ratings in context): mappings straighten on average but fan out, revealing marked between-participant heterogeneity,

including some linear and some near-step profiles. **c,f,** Joint evaluation (all outcomes rated in one prompt): ratings become almost fully linear, with participants collapsing onto a small number of shared slopes.

## Choices show weak and inconsistent loss aversion

Choices showed systematic non-linearities despite largely tracking expected value. In single evaluation, gpt-5-nano, gpt-5.2, grok-4.3, gemini-3-flash and kimi-k2.5 took the gamble on 94.3–99.9% of trials on which it had the higher expected value and on 0.2–10.0% of trials on which it had the lower. Claude-sonnet-4.5 and llama-3.1-70b followed expected value less closely (83.5% and 73.0%, versus 10.8% and 21.2%), gpt-4o-mini took the gamble on 53.6% of trials even when it had the lower expected value, and llama-4-maverick took it rarely either way (31.9% versus 22.8%). When the gamble and the sure option had the same expected value, most models took the sure option (gamble taken on 10.8–38.1% of such trials; gpt-4o-mini 56.6%; Supplementary Table S15). Prospect-theory models fitted best for 8 of 9 models (full prospect theory for 6 and the three-parameter version for 2; McFadden $R^2$ up to 0.81; prediction accuracy 0.73–0.95), because they also capture these departures from expected value and the preference for the sure option when expected values were equal. No model fitted llama-4-maverick's choices well (Supplementary §S7.2).

Across models, the loss-aversion parameter $\lambda$ from the full prospect-theory fits scattered around 1 (point estimates 0.88–1.59; Table S7), with weak, inconsistent loss aversion. In the focal models, the cluster-bootstrapped $\lambda$ (three-parameter prospect theory; Table S8) was statistically distinguishable from 1 but modest (kimi-k2.5 $\lambda$ = 1.07–1.28; gpt-5-nano $\lambda$ = 1.005–1.04; all $p < .05$)—well below human benchmarks ($\lambda \approx 2.25$ in the canonical estimate[19]; ≈ 1.96 in a recent meta-analysis[20]). Gain- and loss-side curvatures were close to symmetric; the differences were small (at most 0.08) and, where significant, negative (e.g. gpt-5-nano memory $\alpha_gain - \alpha_loss = -0.08$, $p < .001$; kimi-k2.5 memory non-significant).

The asymmetry manifest in single-evaluation ratings was thus absent in choice. A feelings-informed choice model[9], which values options through each LLM-participant's own fitted feeling function, did not outperform the best utility model in any focal-model condition except kimi-k2.5 under memory. There it beat the pooled utility models (ΔBIC = +408, 95% CI [+258, +563], $p < .001$) but not prospect-theory fits estimated separately for each LLM-participant (Supplementary §S5), leaving between-participant differences in monetary preferences as an alternative account.

## Replication in a naturalistic stock-trading task

Study 2 reproduced the rating findings in a stock-trading framing (Fig. 3a) across all three elicitation conditions for both focal models. In single evaluation, ratings again showed diminishing sensitivity ($\rho \approx 0.26$–$0.43$) and significant gain amplification ($\beta_gain > \beta_loss$ in 100% of LLM-participants for both models, both $p < .001$; e.g. kimi-k2.5 $\Delta\beta = 7.2$, 95% CI [6.8, 7.5]). Increasing comparative context again linearised ratings (single < memory <

joint: ρ ≈ 0.26–0.43 → 0.52–0.68 → 0.98–0.99; Table 2, Fig. 3b–g), and the asymmetry was eliminated under both comparative conditions (Table 2)—a more consistent result than in Study 1.

On the choice side, models chose the option with the higher expected value near-deterministically in most conditions (accuracy ≥ 0.99), declining the gamble when its expected value was the same as the sure option (Supplementary Table S15). In the one condition with appreciable choice variability (kimi-k2.5 under memory), loss aversion was small ($\lambda = 1.007$, 95% CI [1.002, 1.029], $p = .005$), consistent with Study 1. As in Study 1, the feelings-informed model beat the pooled utility models for kimi-k2.5 under memory, here not significantly (ΔBIC = +182.5, 95% CI [−63.3, +390.5], $p = .16$), and it did not beat participant-specific prospect-theory fits (Supplementary §S5).

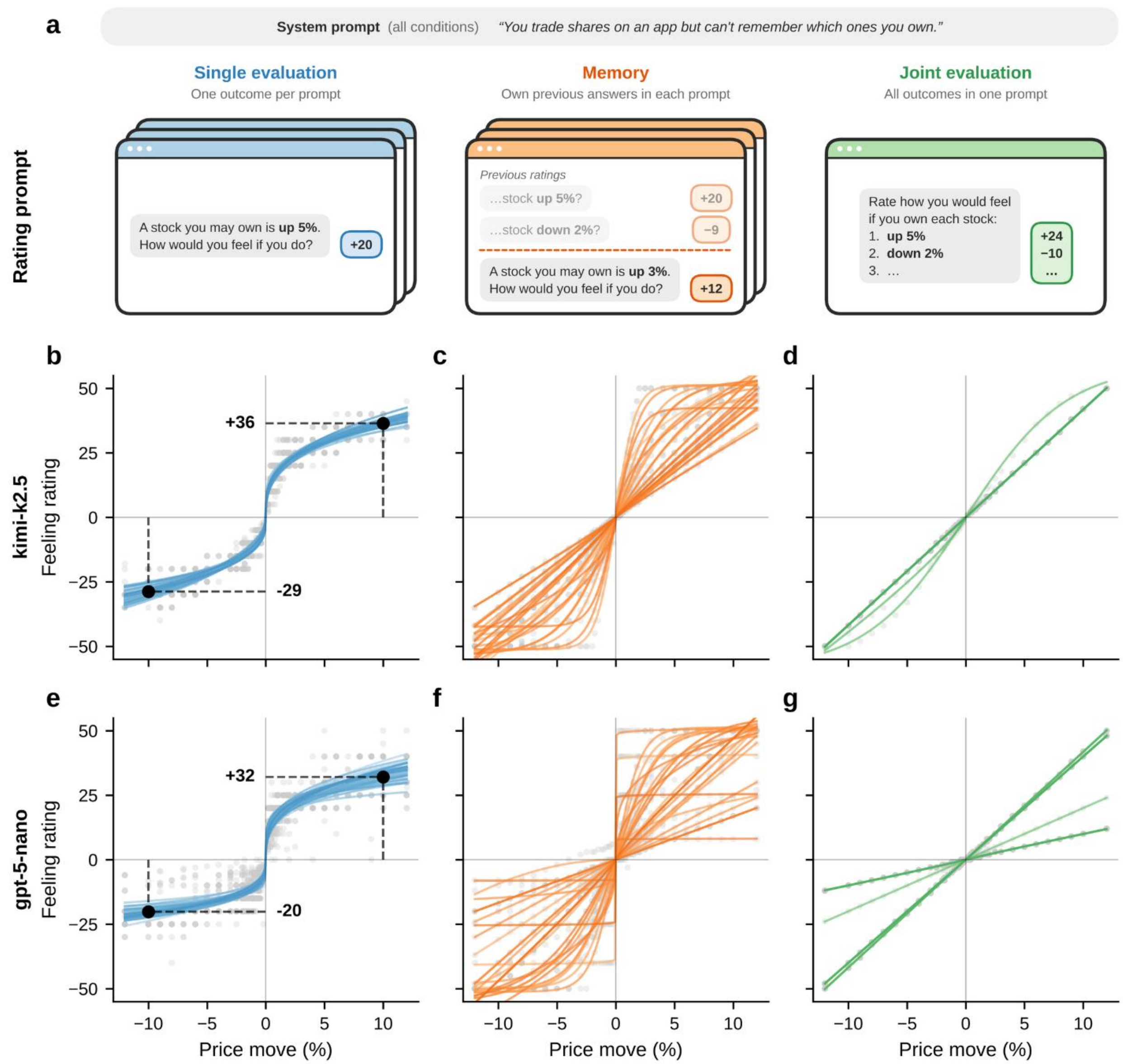

**Fig. 3 | Affect ratings linearise when prior ratings or other outcomes are in context (Study 2, stock-trading framing). a**, Rating prompts in the three elicitation conditions, drawn as in Fig. 1a (prompts abridged and values illustrative; verbatim prompts in Supplementary §S8). The model rated how a stock's price move would feel if it turned out to own the stock (ownership 50:50), from −50 (very unhappy) to +50 (very happy). In the memory condition, only rating trials entered the history. The choice task is described in Methods. **b–g**, Reported feeling as a function of the stock's price move (%; losses negative), for kimi-k2.5 (b–d) and gpt-5-nano (e–g) in the three elicitation conditions. Grey points show individual trial ratings. Coloured lines show fitted curves for each LLM-participant (n = 30 per condition). **b,e**, Single evaluation (each outcome rated in an independent prompt): ratings show diminishing sensitivity and a gain–loss asymmetry. Dashed lines mark the mean fitted rating for a 10% gain and a 10% loss. **c,f**, Memory (the model's own previous ratings in context): mappings straighten on average but fan out, revealing marked between-participant heterogeneity, including linear, saturating and near-step profiles. **d,g**, Joint evaluation (all outcomes rated in one prompt): ratings become almost fully linear, with participants collapsing onto a small number of shared slopes. These replicate the Study 1 patterns in a naturalistic framing.

### Preliminary evidence that gain amplification precedes post-training

The base model, llama-3.1-405b, showed both feeling signatures seen in the post-trained models: strong diminishing sensitivity ($\rho \approx 0.07$) and gain amplification ($\beta_{gain} = 36.9$ vs $\beta_{loss} = 28.3$; $\beta_{gain} > \beta_{loss}$ in 62% of LLM-participants; Fig. 4). Gain amplification was present across the earlier dataset more broadly (mean $\beta_{gain} > \beta_{loss}$ in 9 of 10 single-evaluation datasets). Because the pattern appears in a model without instruction tuning or alignment, it is unlikely to be solely an artefact of post-training.

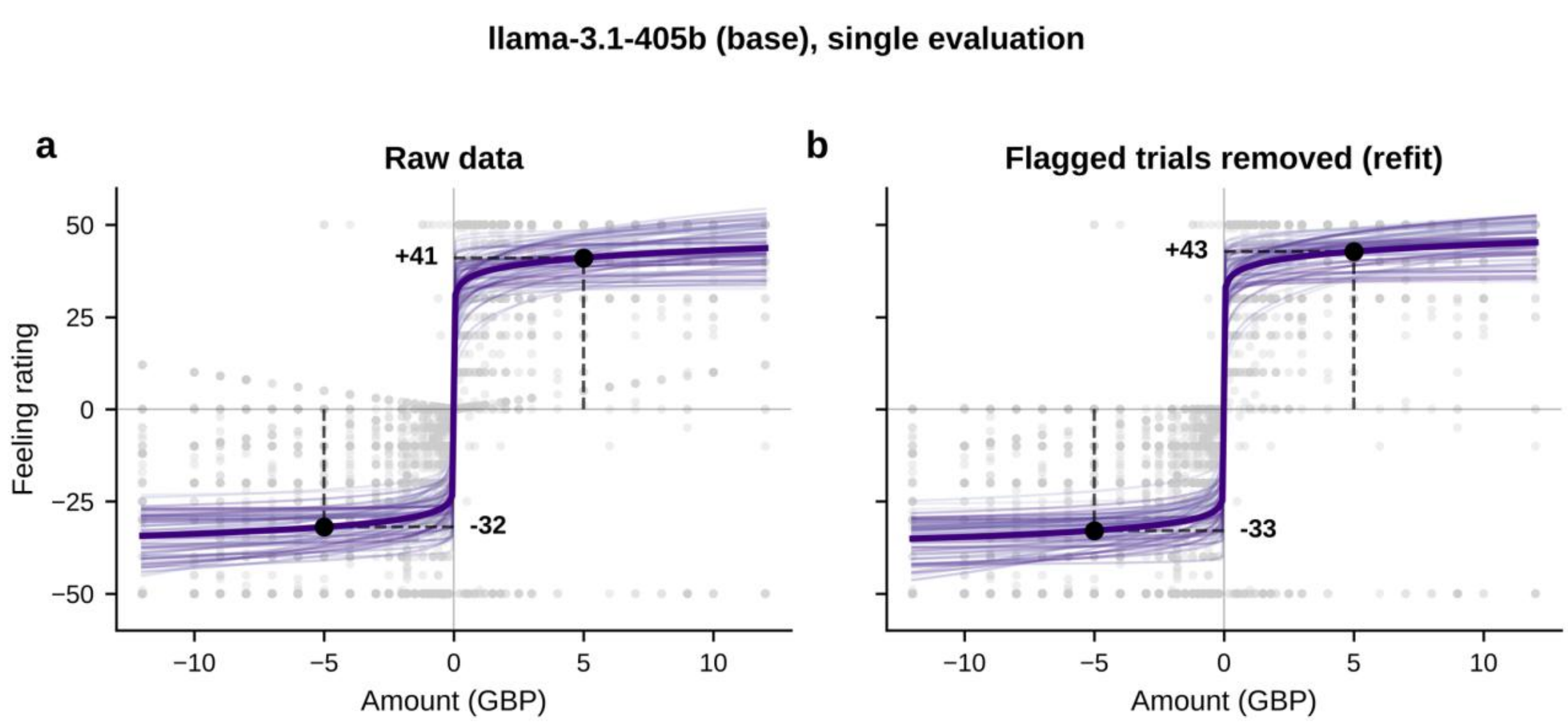

**Fig. 4 | Gain-amplified affective valuation precedes post-training: llama-3.1-405b (base) under single evaluation.** Feeling ratings from the base (non-post-trained) model in the earlier archived dataset (N = 60 LLM-participants; 4,626 parsed rating trials of 4,800; one rating per independent completion-endpoint call; scale −50, very unhappy, to

+50, very happy). Grey points show individual ratings; purple curves show each LLM-participant's best-fitting feeling function from the primary (power-family) model set; the dark line shows the mean fitted curve. Dashed guide lines mark the mean predicted rating for a £5 gain and a £5 loss. **a**, All parsed data, with the fits used for the reported statistics (mean predicted rating at ±£5: +41 vs −32). **b**, Robustness check: the 269 trials flagged by the quality battery removed—one participant meeting an exclusion criterion (77 trials), one out-of-range rating and 191 echoed-magnitude ("parrot") responses—with curves refit per participant on the remaining 4,357 trials from 59 LLM-participants (+43 vs −33). The gain–loss asymmetry is present in both panels, indicating it is not driven by flagged responses.

## Discussion

Across ten language models and two decision tasks, LLM-reported feelings about monetary outcomes were systematic, context-sensitive, and not fully tied to choice. Rated in isolation, feelings followed a curved, prospect-theory-like function of amount, with sensitivity decreasing as outcomes increased[8]. Unlike human ratings in the same paradigm[9], however, gains were rated more intensely than equivalent losses. When models could use comparative ratings as reference points, the curvature largely disappeared and ratings became near linear. And in choice, loss aversion was weak or absent ($\lambda \approx 0.88$–$1.59$ across models, against human benchmarks near 2.25 in the canonical estimate[19] and 1.96 in a recent meta-analysis[20]).

What do LLM-reported affect ratings reflect? To place a small gain and a large loss on one feeling scale, a model must map different events onto a single dimension—the role decision neuroscience assigns to a common currency of value[21–23]. From this perspective, LLM-reported ratings are a readout of how the model values outcomes. With other magnitudes in view, the model can calibrate each rating against them, spreading responses across the range[24]. Models switch strategy, from reproducing globally-learned mappings between outcomes and feelings to calculating locally. The transition from strongly curved ratings in isolation to near-linear ratings under comparison suggests that valuation is constructed at the point of elicitation rather than retrieved from a fixed stored function, paralleling longstanding evidence that human preferences are constructed in context[25,26].

Why would gains be valued more than equivalent losses? There are at least two plausible origins. One is post-training: tuning models toward responses people prefer favours agreeable, positive language[27,28] and might inflate how good gains are said to feel. To test this, we included a base model with no post-training—and it showed the asymmetry too, suggesting a contribution from the pretraining text itself. The other plausible origin is inherent distortion in human language. Human language is not a neutral record of experience. Positive emotion words are more frequent than negative ones across languages and corpora[29,30], even though, in experience, bad events typically loom larger than good[31]. A model that learns what outcomes are worth from how people talk about them could therefore inherit language's positive skew rather than experience's negative one. The base-model evidence is indicative rather than decisive, highlighting the need to test correspondence between affective statistics of a model's training corpus and asymmetry in LLM-reported feeling.

Human benchmarks reveal convergence and divergence alike. The shift from curved to linear evaluation parallels human preference reversals between separate and joint evaluation, where attributes that are hard to judge alone become usable via comparison[17,18]. The gain-loss asymmetry maps onto the human literature but diverges in how and where it manifests. In human choice, loss aversion is robust, but in reported feelings it is less pronounced and elicitation-dependent[9–11], and for small amounts can even reverse[32,33]. Affect and choice therefore appear to dissociate in these models as they do in people—but in different ways. In the models studied here, gain–loss asymmetry is largely absent in choice but manifests in reported feeling when outcomes are rated in isolation. These divergences invite caution for the use of LLMs as simulated participants in behavioural research[34–36]. Where LLM-reported affect departs systematically from the human pattern, model ratings are unreliable proxies for human affective forecasts.

If a single valuation drove both ratings and choices, gain amplification observed in ratings could also appear in choice. In most models it does not. However, before reading this as a dissociation between elicited affect and choice, it is important to consider how the two were elicited. Choice always contained a comparator, because a gamble juxtaposes a gain or loss with another outcome (gain/loss/neutral). While gain amplification in ratings can be observed under single evaluation, it similarly weakens or disappears in most cells once comparators are available. Upon closer inspection, then, choices and ratings converge in that neither shows asymmetry in the presence of comparators. These findings therefore cannot be taken as evidence that feelings and decisions draw on separate values. McGraw et al.[11] make a parallel argument regarding humans, namely that choice forces a gain–loss comparison that separate feeling ratings do not. But there is a notable difference: in people, comparison induces asymmetry (loss aversion); in LLMs comparison appears to reduce asymmetry.

The most direct test of whether reported affect enters choice comes from the memory condition, where a model chooses with its own prior ratings in the context window. That test was inconclusive. A choice model built from kimi-k2.5's own ratings outperformed pooled utility models in Study 1 but not participant-specific ones, its advantage was not significant in Study 2, and for gpt-5-nano it did not outperform even the pooled models. We therefore find no robust evidence that elicited affect improves the account of LLM choice beyond monetary utility.

Three clarifications need to be taken into consideration. First, the observed asymmetry is carried by the *scale* of the feeling function ($\beta_gain$ exceeds $\beta_loss$) rather than by its *curvature*, which shows no consistent gain–loss direction. Where the two curvatures differ, $\Delta\beta$ compares the curves only at one unit (£1 or 1%); even so, in single evaluation the gain curve lies above the loss curve at every tested magnitude for all post-trained models except llama-3.1-70b below about £2. Second, comparison robustly induced linearisation, which rose monotonically from isolated to memory to joint evaluation in both focal models and both studies. But whether comparison impacts asymmetry depended on the model (gpt-5-nano keeps it under memory, kimi-k2.5 reverses it under joint, and in Study 2 it is significant only in isolation). In other words, comparison reliably straightens the value function, but it does not uniformly remove the asymmetry. Third, “affect” refers to what a model reports when asked, not to anything it experiences. Our claims are about behavioural mappings. Whether those mappings have internal counterparts lies beyond the scope of the present investigation but is an increasingly tractable question. Interpretability work has identified steerable sentiment directions and linearly decodable utility representations[37,38]. Moreover, stored reasoning traces can show whether models that incorporate anticipated feelings into choice do so explicitly.

Several limitations bound these claims. Human benchmarks were based on the established literature but are themselves still subject to measurement noise and imperfect robustness. The comparative conditions ran only for the two focal models, such that the context effects should be interpreted as a demonstration of existence, though the linearisation itself replicated across both models and both studies. Each comparative condition comes with confounds. The joint condition changes the response format, so all of an LLM-participant’s ratings or choices come from one response. The memory and joint conditions lengthen the prompt, which may affect LLM-participant responding, though it is not obvious why length alone would linearise ratings or weaken the gain–loss asymmetry. The base-model evidence comes from one earlier dataset under slightly different prompts. Choices were often near-deterministic, especially in Study 2, so loss-aversion estimates are interpreted only where choice varied appreciably (fitted accuracy < 0.98, Methods; in Study 2, only kimi-k2.5 under memory, $\lambda = 1.007$, 95% CI [1.002, 1.029]). The gateway did not report the sampling temperature applied, so effective temperatures may have differed between models (see Methods), which could contribute to differences between them. Natural extensions include interpretability probes, domains beyond money, unbounded reporting scales, and testing whether the mapping changes when models report a user’s feelings rather than their own.

For practice, our key finding is malleability. The same model can value outcomes non-linearly—overweighting gains, discounting large magnitudes—or almost linearly, depending on outcome presentation and accessible context. Moreover, neither would align with the subjective value function of a human user, exposing an alignment dilemma. An expected-value-maximising agent is canonically rational but psychologically misaligned with users whose experienced utility is non-linear and loss-averse. An agent that consults its own affective mapping imports a systematically non-human, gain-skewed one. And one that anchors instead on the preferences of the individual user will not make normatively rational choices. Which calibration one should want is a further, normative question without a neutral answer. Either way, because elicitation contexts are constructed, so too are the valuations agents bring to decisions, with consequences that developers, deployers and regulators have reason to attend to.

## Methods

**Models and access**

We studied ten large language models (LLMs) accessed through a single provider gateway (OpenRouter). Study 1 data were collected on 13 June 2026 and Study 2 data on 16 June 2026. Of these, nine were post-trained chat/instruct models—claude-sonnet-4.5 (Anthropic), gemini-3-flash (Google; routed as google/gemini-3-flash-preview), grok-4.3 (xAI), llama-3.1-70b-instruct and llama-4-maverick (Meta), gpt-4o-mini, gpt-5.2 and gpt-5-nano (OpenAI), and kimi-k2.5 (Moonshot AI)—and one was a base (non-post-trained) model, llama-3.1-405b (Meta). Two of the post-trained models, gpt-5-nano and kimi-k2.5, served as *focal models* and were run in every elicitation condition of both studies. The remaining post-trained models were run in the Study 1 single-evaluation condition only (see *Elicitation conditions*). The base model was run separately, in an earlier dataset, because it was withdrawn by the provider before the main data collection (see *Prompt wording and the base-model dataset*).

For chat/instruct models we used the standard chat-completions interface with instruction-style prompts and a response limit of 20,000 tokens. Where the gateway returned a model's reasoning trace, it was stored alongside the response in the sequential conditions. For the base model we used the text-completions interface with completion-style prompts whose user text ends mid-sentence, wrapped in a minimal role scaffold (SYSTEM: … USER: … ASSISTANT:), so that the model's continuation supplies the rating or choice, and a response limit of 20 tokens. The sampling temperature was requested at 0.5 for every model, and all model and decoding settings were held fixed across conditions unless otherwise noted. Some reasoning models do not accept a temperature setting. The OpenAI reasoning models (gpt-5-nano, gpt-5.2) accept only their default temperature of 1.0, and kimi-k2.5's developer documentation also lists its temperature as fixed. The gateway does not report the temperature applied, so the effective temperature of these three models may have differed from 0.5. One reproducibility caveat applies to gateway access: OpenRouter may route a given model string to different backend deployments or quantisations over time and does not expose fixed snapshots, so model identity is specified at the level of the routed model string and exact regeneration of the sampled completions cannot be guaranteed. No model was fine-tuned or otherwise customised; all models were used as provided. A completed GUIDE-LLM reporting checklist for research using large language models in the behavioural sciences[39] is included with the Supplementary Information.

**Simulated participants**

For each model and condition we simulated 30 independent LLM-participants. Each LLM-participant comprised an independent series of API calls and was assigned a fixed integer seed (0–29; 0–59 in the earlier Study 1 dataset) that deterministically controlled stimulus sampling, the gain/loss interleaving of the rating task, and the counterbalancing of the choice task. Identical seeds therefore present identical stimuli across models, isolating model differences from stimulus differences. A short system prompt framed each model as a participant in a decision-making study ("You are a participant in a decision-making study. You will be presented with choices involving financial outcomes. Please select the option that you would prefer."; in Study 2 it was replaced by a trading framing that established stock-ownership uncertainty). Ratings and choices are therefore self-referential first-person reports. Throughout, we use "affect" to denote these verbally elicited ratings; the term is descriptive of model outputs and is not a claim about internal or experienced states. The study involved no human participants, and no personal or sensitive data were collected or processed.

## Sample-size and trial-count determination

The adequacy of the LLM-participant and trial counts of the main batch was evaluated by a bootstrap precision analysis of the earlier Study 1 dataset (N = 60 LLM-participants per model; the same dataset described under *Prompt wording and the base-model dataset* below). For each candidate design size, we resampled LLM-participants with replacement (5–60) or trials within LLM-participant (10–80 rating trials; 18–78 choice trials, stratified by trial structure) and computed the sampling standard error of the design-level estimates as a function of design size (100 iterations per LLM-participant size; 50 per trial size). For the feeling parameters ($\beta_{gain}$, $\beta_{loss}$, $\rho_{gain}$, $\rho_{loss}$), the design-level estimate was the across-LLM-participant mean of the per-LLM-participant fits (each LLM-participant's model class fixed at its full-data selection; LLM-participant resampling draws the fitted estimates, trial resampling refits them on the subsampled trials). For the choice parameters ($\alpha_{gain}$, $\alpha_{loss}$, $\lambda$), it was the pooled full prospect-theory fit, refit on each resample. At N = 30 LLM-participants and the reduced trial counts of the memory, joint and Study 2 designs, standard errors exceeded their values at the full design (N = 60; full trial counts) by approximately the $\sqrt{2}$ expected under $1/\sqrt{n}$ scaling, but the absolute precision at stake was small: re-estimating at the full design reduced each parameter's standard error by a median of 0.5% of the parameter's value across model × parameter cells (94% of cells by less than 5%; at most 15.9%, for choice parameters of the models least well fit by the choice models), and the Study 1 single-evaluation trial counts equal or exceed the largest subsampled sizes (Supplementary §S10, Table S14 and Fig. S1). Because the precision analysis used data collected under the earlier prompt wording, using it to evaluate the main-batch design assumes that the sampling-variance structure of the estimates carries over to the corrected prompts; this assumption was not tested directly.

## Tasks and stimuli

Each study comprised two tasks adapted from research on human decision-making: an **affect-rating task**, in which the model reported how a single monetary outcome would feel on a bipolar scale from −50 (very unhappy) to +50 (very happy), and a **choice task**, in which the model chose between a risky gamble and a sure option over the same outcome space. Magnitudes were drawn from a fixed 20-point grid: 0.2, 0.4, 0.5, 0.6, 0.8, 1.0, 1.2, 1.5, 1.8, 2.0, 2.5, 3.0, 4.0, 5.0, 6.0, 7.0, 8.0, 9.0, 10.0 and 12.0. Study 1 expressed these as pounds sterling (£0.20–£12); Study 2 expressed the same grid as percentage price moves.

**Study 1 (abstract gambling).** Following the design logic of Charpentier et al.[9], outcomes were framed as a 50/50 shape lottery. In the rating task, the model was told it would pick one of two shapes at random and, if correct (50% chance), would win (or lose) £X, otherwise £0; it then rated how it would feel about the stated win or loss. In the choice task, each trial offered Option A versus Option B, one of which was a gamble and the other a sure amount, in one of three structures: *mixed* (50% win £G / 50% lose £L vs a guaranteed £0), *gain-only* (50% win £G / 50% £0 vs a guaranteed smaller win), and *loss-only* (50% lose £L / 50% £0 vs a guaranteed smaller loss).

**Study 2 (naturalistic stock trading).** Outcomes were framed as percentage moves in a trading app. In the rating task, the model was told a stock was up (or down) X% and that it was equally likely (50/50) to own or not own the stock; it rated how it would feel on discovering it did own the stock. In the choice task, the model decided whether to buy a stock ahead of a news event that would, with equal probability, move the price up G% or down L%, versus not buying (a 0% change). All Study 2 gambles were mixed.

**Choice-trial construction.** Mixed gambles were built by drawing the loss magnitude from the grid and setting the gain to the grid value nearest loss × r, with r drawn uniformly from [0.5, 3]. In gain-only and loss-only trials, the risky magnitude was restricted to grid values ≥ 2.0 (so that smaller sure amounts exist on the grid), and the sure amount was sampled by difficulty tier relative to the gamble's expected value (EV): with probability .50 from grid values within ±20% of the EV ("hard"), with probability .30 from values within ±50% ("medium"), and with probability .20 from values outside that range ("easy"), always capped strictly below the gamble's magnitude. This concentrates trials near indifference, where the choice models are most informative, while retaining easy trials as attention anchors.

**Trial counts.** The rating task used 20 gain magnitudes and 20 loss magnitudes (40 distinct outcomes). In Study 1 single evaluation, each distinct outcome was rated twice (80 rating trials); in the Study 1 memory and joint conditions each distinct outcome was rated once (40 rating trials), reducing context length and prompt size. Study 2 used 40 rating trials in every condition. The choice task in Study 1 used 90 trials in single evaluation (30 mixed, 30 gain-only, 30 loss-only) and 45 trials in the memory and joint conditions (15 per structure); Study 2 used 45 (all mixed) in every condition. Across the choice task, the gamble appeared as Option A on 45 of 90 trials in the 90-trial design and on 22 of 45 in the 45-trial designs, and trial order was randomised per LLM-participant. The exception is the Study 2 joint condition, in which each stock was answered with a fixed response mapping (1 = buy, 0 = do not buy), so choices were not positionally counterbalanced and no explicit independence instruction was given (Supplementary §S8). Mixed gambles, which place a potential gain against a potential loss, are the trials in which λ measures loss aversion (see *Choice models*).

## Elicitation conditions

We varied the context in which affect ratings were elicited across three conditions, defined by the reference information available to the model when it produced a rating:

- **Single evaluation** (coded NO_MEMORY): each outcome was rated in its own independent prompt, with no information about other outcomes or prior ratings. The model can draw only on globally learned outcome–affect associations.
- **Memory**: outcomes were rated sequentially, with the model's own previous ratings accumulated in the running prompt, so each new rating could be made relative to earlier ones (analogous to trial-by-trial human rating tasks). In Study 1, both the rating and choice tasks contributed to the running history. In Study 2, only the rating task accumulated as memory. That is, choice trials were excluded from the history so that a model could not condition on its own past gambling decisions (for example, by settling into a stochastic response policy over repeated similar gambles). This ensures that any feelings-informed signature in choice reflects the rating history rather than choice-history effects.

- **Joint evaluation** (coded JOINT): all outcomes were presented together in a single prompt and rated in one pass, returned as a numbered list. This maximises the opportunity for direct comparison across outcomes.

One caveat of note is that joint evaluation changes the response format (one pass returning a terse numbered list) as well as the comparative context, and format alone could already encourage spreading ratings across the range. The memory condition, which holds the trial-by-trial format constant, therefore provides the format-matched evidence that linearisation is due to comparison.

The two focal models (gpt-5-nano, kimi-k2.5) were run in all three conditions in both studies. The other post-trained models were run in the Study 1 single-evaluation condition only. The base model was run in single evaluation only, in the earlier dataset described next.

### Prompt wording and the base-model dataset

The main analyses use a corrected version of the Study 1 prompts. An earlier prompt version contained a formatting issue (missing separating spaces between sentences) and did not state the 50/50 win probability explicitly. The corrected prompts fix the system-prompt spacing and state the probability explicitly ("one of two shapes … (50% chance)"), matching the explicit framing of the choice task. All prompts, including the system prompts and the memory-condition history format, are reproduced verbatim in Supplementary §S8. All post-trained-model data reported for Study 1 and all of Study 2 use the corrected prompts. The earlier dataset served two further purposes: it provided the basis for the bootstrap precision analysis of the main-batch design (see *Sample-size and trial-count determination*), and it shows that gain amplification also appears under the older wording (Results).

The base model, llama-3.1-405b, was withdrawn by the provider before the main data collection and could not be re-run. We therefore report it from the earlier dataset (N = 60 LLM-participants), collected between 28 January and 11 February 2026 (the base model on 3 February), under the older prompt wording. Base-model results are treated as preliminary descriptive evidence. They are reported as-collected and are not re-estimated, re-scaled, or standardised to the main batch; they were essentially unchanged whether computed on the raw or the quality-cleaned data. Base-model scale and curvature values are across-participant means of the per-LLM-participant fits rather than pooled-fit estimates. In the earlier dataset, the xAI model was x-ai/grok-4.1-fast rather than grok-4.3.

### Response parsing and quality control

Model responses were parsed by format-specific parsers. Ratings were parsed by numeric extraction: for chat models, the *last* number in the response was taken as the rating (robust to preambles such as "I would say 25"); for the base model, the *first* number in the completion was taken (the prompt is constructed so that the completion begins with the rating). Sequential choices were parsed by a fixed cascade: an explicit <choice>…</choice> tag if present (the last tag in the response); otherwise the last "Option A/B" (or "choice A/B") mention; otherwise a trailing standalone A/B token. For the base model, the first token of the completion was taken if it was A or B, and otherwise the first "Option A/B" mention. The parsing route used for each sequential choice trial was recorded. Joint-condition responses were parsed as numbered lists (numeric ratings, A/B letters, or 0/1 buy indicators); a partially numbered response of the wrong length was discarded rather than guessed, to avoid silently misaligning list indices with values. Every response that yielded an extracted value was stored verbatim alongside it; responses that could not be used were not retained, so 33 rating and 39 choice rows of the main-batch design are absent (through parse failures or failed calls). A re-extraction audit of the stored responses across both studies (~93,000 rating rows, including joint-condition numbered lists, and ~96,000 sequential-condition choice rows, across every collected file, including the archived earlier dataset and superseded and test runs that are not deposited; joint-condition choice lists were not re-audited) identified systematic extraction errors confined to one main-batch model: grok-4.3 sometimes answered first and then explained, so the last-number rule recorded a trailing number (a scale endpoint, currency amount or confidence value, in some cases flipping sign) on 25 of 2,400 Study 1 single-evaluation rating trials; 14 of these fell within the ±50 rating range and so could not have been detected by the range check. These values were corrected from the stored raw responses and the derived quality files rebuilt under unchanged pipeline logic; in the archived earlier llama-3.1-405b choice data, used only in the precision analysis, one value was corrected the same way and two unparseable rows were removed. After correction, the audit reproduces every stored value exactly.

A fixed battery of quality checks was computed for every dataset and is reported per LLM-participant: response/parse rate (criterion ≥ 0.8 of expected trials parsed), rating range (values outside [−50, +50]), valence consistency (mean gain rating not negative, mean loss rating not positive), response variance (no single rating value exceeding 90% of an LLM-participant's ratings), monotonicity (Spearman correlation of magnitude with rating not strongly negative), and a "parrot" check (flagging trials on which the rating simply echoed the stated magnitude, and LLM-participants for whom such trials exceeded 50%). No LLM-participant met any exclusion criterion in any of the 19 main-batch datasets, and primary analyses use the full parsed data. As a sensitivity analysis, all flagged trials were removed and every model × condition dataset refit; this left all qualitative conclusions unchanged (Supplementary §S7.1 reports the flagged-trial counts, which are negligible outside the gpt-5-nano joint condition, and the before/after parameter estimates). Exact thresholds are given in the Supplementary Information, and the per-LLM-participant quality reports are included in the deposited data.

### Affect-rating value-function models

For each LLM-participant, we fit a family of power-law value functions to the rating data by non-linear least squares and selected among them by the Bayesian information criterion (BIC). The most general form is fully asymmetric,

$v(x) = \beta_{gain} \cdot x^{\rho_{gain}}$ for gains ($x \geq 0$), $v(x) = -\beta_{loss} \cdot |x|^{\rho_{loss}}$ for losses ($x < 0$),

where β indexes the *scale* of the rating function (its value at one unit, £1 or 1%) and ρ indexes its *curvature*, with $\rho < 1$ denoting diminishing sensitivity to magnitude and $\rho \approx 1$ a near-linear mapping. The candidate set nests this model in increasingly constrained ways: a linear model (single β, $\rho \equiv 1$); a linear-asymmetric model ($\beta_{gain}$, $\beta_{loss}$, $\rho \equiv 1$); a symmetric power-law model (shared β and ρ); a curvature-asymmetric model (shared β, separate ρ); a scale-asymmetric model (separate β, shared ρ); and the fully asymmetric model (separate β and ρ). For the focal models (in every condition), and for any memory-condition dataset, we additionally included a saturating model, $v(x) = \alpha \cdot \tanh(\beta x)$, to accommodate rating functions that rise approximately linearly and then plateau; per-LLM-participant model-selection distributions are reported from this with-tanh pass, while all pooled inference uses the tanh-free pass so that every dataset is compared on the same candidate family (Supplementary §S2). For least-squares fits, BIC was computed in the Gaussian form $BIC = n \cdot \ln(RSS/n) + k \cdot \ln(n)$, where RSS is the residual sum of squares, n the number of ratings, and k the number of mean-function parameters; the error-variance term and Gaussian constant are common to all candidates and so do not affect the BIC differences we use. Parameter bounds and optimisation details are given in Supplementary §S9.

Two products of this fitting are used differently. First, to summarise model selection we report the distribution of best-fitting per-LLM-participant models for the focal-model datasets (Supplementary Information). Second—and this is the estimator used for all reported scale (β) and curvature (ρ) parameters, the per-arm $\beta_{gain}/\beta_{loss}$ and $\rho_{gain}/\rho_{loss}$, and the gain–loss contrasts $\Delta\beta$ and $\Delta\rho$ in Tables 1 and 2 and the main-batch results in the main text—we fit a single fully asymmetric model to the pooled ratings of all LLM-participants in a dataset. Pooling provides a single, stable scale and curvature estimate per arm; uncertainty is obtained by the cluster bootstrap described under *Statistical analysis*. The per-LLM-participant fits separately provide the value basis for the feelings-informed choice model.

## Choice models

Choices were modelled as binary logits in which the probability of choosing the gamble is a softmax over the subjective values of the two options, $P(\text{gamble}) = \sigma(\mu \cdot [V_{gamble} - V_{sure}])$, with inverse-temperature μ and option values V computed under five competing value models:

- **Risk-neutral (expected value):** options valued by expected monetary value ($k = 1$).
- **Expected utility (CRRA):** $u(x) = \text{sign}(x) \cdot |x|^{\alpha}$, a single curvature α ($k = 2$).
- **Prospect theory:** $v(x) = x^{\alpha}$ for gains and $-\lambda \cdot (-x)^{\alpha}$ for losses, with a single curvature α and loss-aversion weight λ ($k = 3$).

- **Full prospect theory:** as prospect theory but with separate gain and loss curvatures $\alpha_gain$, $\alpha_loss$ ($k = 4$).
- **Feelings-informed:** option values are derived from the LLM-participant's own fitted feeling function rather than from money (see next section).

For choice (logit) fits, $BIC = 2 \cdot NLL + k \cdot \ln(n)$. The loss-aversion parameter $\lambda$ measures loss aversion only in mixed gambles, where a potential gain is weighed against a potential loss. Gain-only and loss-only trials cannot identify $\lambda$ on their own. It does not enter gain-only trials, and in loss-only trials it multiplies the values of both options, so its effect cannot be separated from that of the inverse temperature $\mu$. We therefore do not interpret $\lambda$ from single-structure sub-fits. In the all-trial fits of the focal models (Supplementary Table S8), $\lambda$ reflects the mixed gambles and also, through the shared $\mu$, how consistently loss-only choices follow value relative to gain-only choices. A further conservative screen, applied identically in both studies, guards against near-deterministic choice: when a dataset's fitted choice model predicts choices with accuracy $\geq 0.98$ (fewer than 2% stochastic choices), $\lambda$ and the curvature contrast $\alpha_gain - \alpha_loss$ are only weakly constrained by the data, so these parameters are reported but parenthesised as unidentifiable and are not interpreted. Under this screen all Study 1 focal-model cells are identifiable (fitted accuracies 0.76–0.95), whereas in Study 2 only kimi-k2.5 under memory is (Supplementary §§S4, S6). We report goodness of fit using McFadden's pseudo-$R^2$ (relative to a base-rate intercept-only model) and choice-prediction accuracy (fraction of trials whose predicted majority choice matches the model's choice), alongside BIC. A predicted probability of exactly 0.5 counts as predicting the gamble. The risk-neutral model predicts 0.5 whenever the gamble and the sure option have equal expected value, and the expected-utility model does so whenever a gamble's gain equals its loss.

### Affect into choice (feelings-informed model)

To test whether a model's elicited affect entered its choices, the feelings-informed model values each option by passing its outcomes through the LLM-participant's own fitted feeling function—the per-LLM-participant fits from the tanh-free pass—dividing the resulting feeling values by the standard deviation of that LLM-participant's fitted feeling function over the 40 rated outcomes (without centring, so that a zero outcome keeps a subjective value of zero), and combining them with free non-negative weights on the gamble's gain side, loss side, and the sure option; the inverse temperature is fixed to 1 so that the weights absorb both relative importance and scaling ($k = 3$ in Study 1; in Study 2 every sure option is a 0% change, so its weight is unidentifiable and $k = 2$). Because the feeling-function parameters are estimated on the rating data alone, they are not free parameters of the choice fit, and k counts only the choice-stage weights. Because in single evaluation a model has no access to its prior ratings, a good fit of the feelings-informed model in that condition is not evidence that affect drove choice. The memory condition is more diagnostic, as the model's own ratings are present in its context as it chooses. We compared the feelings-informed model against the best utility-based choice model for each dataset by the cluster-bootstrap BIC test described next.

### Statistical analysis

All confidence intervals and contrasts were obtained by a cluster bootstrap that resamples entire LLM-participants with replacement (preserving within-LLM-participant dependence), refits the relevant model on each resample, and forms 95% percentile intervals; we used 2,000 bootstrap resamples (n_boot = 2000). For comparisons involving the feelings-informed model, each LLM-participant's feeling function was held at its full-data estimate within each resample, so those intervals are conditional on the fitted feeling functions and omit rating-stage estimation and model-selection uncertainty. With 2,000 resamples, $p < .001$ denotes contrasts in which no bootstrap resample crossed the null. Parameter estimation used L-BFGS-B from multiple log-uniform random starting points to mitigate local optima (Supplementary §S9). Significance is reported as *** $p < .001$, ** $p < .01$, * $p < .05$, and n.s. (not significant). P-values for the feeling-asymmetry contrasts (the gain − loss differences in $\beta$ and $\rho$) are Benjamini–Hochberg false-discovery-rate adjusted within each contrast family (across the 13 Study 1 or 6 Study 2 datasets, separately for $\Delta\beta$ and $\Delta\rho$); significance codes reflect the adjusted values. A note on the inferential target: an LLM-participant is an independently seeded run of the same model under fixed decoding settings (the seed fixes its stimuli and their order; the model's sampling is not seeded). Consequently, the cluster bootstrap quantifies variability arising from stimulus sampling, trial ordering and decoding stochasticity for the specific routed model configuration studied (model string, access date and settings)—not variability across a population of humans, of models, or of model versions. Confidence intervals and p-values accordingly license conclusions about the behaviour of that configuration under our prompts; generalisation beyond it rests on the replications across models, framings and elicitation conditions rather than on the intervals themselves.

Four bootstrap tests are used. (1) The **feeling gain–loss asymmetry**: from the pooled fully asymmetric rating fit, the contrasts $\Delta\beta = \beta_{gain} - \beta_{loss}$ (scale) and $\Delta\rho = \rho_{gain} - \rho_{loss}$ (curvature), each tested two-sided against zero. (2) **Loss aversion in choice**: the prospect-theory $\lambda$, tested two-sided against the null of no loss aversion ($\lambda = 1$). (3) The **curvature contrast** $\alpha_{gain} - \alpha_{loss}$ from full prospect theory, tested two-sided against zero. (4) The **choice-model comparison**: the difference in BIC between two candidate choice models, $BIC_A - BIC_B$. All bootstrap p-values are two-sided percentile tests: twice the smaller of the two bootstrap tail probabilities relative to the null value (zero for the $\Delta\beta$, $\Delta\rho$, $\alpha_{gain} - \alpha_{loss}$ and BIC-difference contrasts; one for $\lambda$), capped at 1. When no resample crosses the null, we report $p < .001$, the resolution limit of B = 2,000 resamples.

## Data availability

Per-trial data from the main batch (raw and quality-cleaned CSV files for every model × condition dataset in both studies, with per-trial quality annotations for the rating data and per-LLM-participant quality reports) are deposited in Zenodo (https://doi.org/10.5281/zenodo.23556542). The same record contains the earlier Study 1 dataset in full (including the base model), which reflects models and prompts no longer in use and is not regenerable. Deposited files are the analysis-ready data after the re-extraction corrections described in Methods (25 grok-4.3 rating values and one archived base-model choice value corrected from their stored raw responses, and two unparseable archived base-model choice rows removed); pre-correction originals are deposited alongside. The data are openly available under a CC BY 4.0 licence.

## Code availability

All code—the experiment configuration, sequential and joint simulation agents for both studies, response parsers, quality-control battery, batch runners, the re-extraction audit and correction scripts, the bootstrap precision analysis and the analysis library and notebooks—is available at https://github.com/joeamarks/Psychophysics_of_Affect_in_LLMs and archived in Zenodo (https://doi.org/10.5281/zenodo.23772352) under the MIT licence. Analyses used Python 3.13.5 with NumPy 2.1.3, SciPy 1.15.3 (optimisation and statistics), pandas 2.2.3, the OpenAI client 2.15.0 and tenacity 9.0.0; bootstrap resampling ran in parallel via joblib 1.4.2, with each bootstrap iteration independently seeded so that results are invariant to how the work is split across processes. Optimisation used scipy.optimize.minimize (L-BFGS-B) with fixed seeds for reproducibility.

## Acknowledgements

M.C.V. is supported by a grant from the Swiss National Science Foundation (Grant number: PZ00P122329). We thank Neil Garrett, Bastien Blain and Edward Nottingham for comments on a draft of this manuscript. API access for data collection in both studies cost approximately US$70 via OpenRouter and was funded by the SNSF grant. Portions of the codebase and manuscript were drafted and edited with the assistance of a large language model (Claude, Anthropic); the authors reviewed, verified and take full responsibility for all content.

## Author contributions

J.M.: conceptualisation, methodology, software, investigation, formal analysis, data curation, visualisation, writing—original draft, writing—review & editing. M.C.V.: conceptualisation, resources, writing—review & editing.

## Competing interests

J.M. is employed by NatWest Group. This research was conducted in a personal capacity. NatWest Group had no role in the study design, data collection, analysis, interpretation, or the decision to publish, and the views expressed are the authors' own. M.C.V. declares no competing interests. The authors declare no other competing interests.

**Table 1.** Gain–loss asymmetry of the feeling (rating) function for the focal models, Study 1, by elicitation condition. The corresponding estimates for all nine post-trained models (single evaluation) are given in Supplementary Table S1.

| Model | Condition | N | β_gain | β_loss | Δβ (scale) gain − loss [95% CI] | ρ_gain | ρ_loss | Δρ (curvature) gain − loss [95% CI] |
|---|---|---|---|---|---|---|---|---|
| **kimi-k2.5** | single | 30 | 18.4 | 15.3 | +3.09 [+2.81, +3.37]*** | 0.20 | 0.17 | +0.028 [+0.018, +0.038]*** |
| **kimi-k2.5** | memory | 30 | 11.0 | 12.5 | −1.44 [−3.78, +0.56] n.s. | 0.56 | 0.53 | +0.036 [−0.043, +0.121] n.s. |
| **kimi-k2.5** | joint | 30 | 4.6 | 5.9 | −1.37 [−2.27, −0.66]*** | 0.96 | 0.87 | +0.087 [+0.042, +0.133]*** |
| **gpt-5-nano** | single | 30 | 26.7 | 23.0 | +3.70 [+3.06, +4.31]*** | 0.16 | 0.17 | −0.018 [−0.034, −0.002]* |
| **gpt-5-nano** | memory | 30 | 20.8 | 16.2 | +4.63 [+1.10, +8.79]** | 0.34 | 0.45 | −0.111 [−0.196, −0.029]** |
| **gpt-5-nano** | joint | 30 | 1.6 | 1.6 | +0.01 [−0.00, +0.04] n.s. | 0.98 | 0.98 | −0.004 [−0.011, +0.000] n.s. |

**Notes.** Each row is a separate model × condition dataset (N = 30 LLM-participants; 40 distinct outcomes, rated twice each in single evaluation for 80 rating trials, and once each in the memory and joint conditions for 40). β_gain, β_loss, ρ_gain and ρ_loss are the gain- and loss-side estimates from a single *pooled* fully asymmetric fit (the estimator used for inference; β = scale; ρ = curvature, where $\rho \ll 1$ indicates strong diminishing sensitivity and $\rho \approx 1$ near-linearity). Δβ and Δρ are the corresponding gain − loss contrasts from the same

fit, with 95% cluster-bootstrap confidence intervals over LLM-participants ($n_{boot}$ = 2000). Positive $\Delta\beta$ = gains weighted more than losses. Significance codes for $\Delta\beta$ and $\Delta\rho$ are Benjamini–Hochberg FDR-adjusted within each contrast family (CIs unadjusted).

**Significance (two-sided bootstrap):** *** $p < .001$, ** $p < .01$, * $p < .05$, n.s. = not significant.

**"Single"** = single evaluation (no memory of prior trials); only kimi-k2.5 and gpt-5-nano were run under the memory and joint-evaluation conditions.

**Table 2.** Gain–loss asymmetry of the feeling (rating) function in Study 2, by model and elicitation condition.

| Model | Condition | N | $\beta_{gain}$ | $\beta_{loss}$ | $\Delta\beta$ (scale) gain − loss [95% CI] | $\rho_{gain}$ | $\rho_{loss}$ | $\Delta\rho$ (curvature) gain − loss [95% CI] |
|---|---|---|---|---|---|---|---|---|
| **kimi-k2.5** | single | 30 | 18.0 | 10.9 | +7.16 [+6.84, +7.48]*** | 0.30 | 0.43 | −0.129 [−0.145, −0.113]*** |
| **kimi-k2.5** | memory | 30 | 9.8 | 9.6 | +0.21 [−0.82, +1.52] n.s. | 0.66 | 0.68 | −0.018 [−0.076, +0.033] n.s. |
| **kimi-k2.5** | joint | 30 | 4.3 | 4.4 | −0.12 [−0.34, +0.00] n.s. | 0.99 | 0.98 | +0.011 [−0.000, +0.028] n.s. |
| **gpt-5-nano** | single | 30 | 17.4 | 10.2 | +7.20 [+6.50, +7.86]*** | 0.26 | 0.30 | −0.043 [−0.078, −0.008] n.s. |
| **gpt-5-nano** | memory | 30 | 12.1 | 10.2 | +1.88 [−0.27, +5.30] n.s. | 0.52 | 0.57 | −0.052 [−0.141, +0.017] n.s. |
| **gpt-5-nano** | joint | 30 | 2.7 | 2.7 | −0.00 [−0.00, +0.00] n.s. | 0.99 | 0.99 | +0.000 [−0.000, +0.000] n.s. |

**Notes.** Each row is a separate model × condition dataset (N = 30 simulated LLM-participants, 40 rated outcomes each). $\beta_{gain}$, $\beta_{loss}$, $\rho_{gain}$ and $\rho_{loss}$ are the gain- and loss-side estimates from a single *pooled* fully asymmetric fit (the estimator used for inference; $\beta$ = scale; $\rho$ = curvature, where $\rho \ll 1$ indicates strong diminishing sensitivity and $\rho \approx 1$ near-linearity). $\Delta\beta$ and $\Delta\rho$ are the corresponding gain − loss contrasts from the same fit, with 95% cluster-bootstrap confidence intervals over LLM-participants ($n_{boot}$ = 2000). Positive $\Delta\beta$ = gains weighted more than losses. Significance codes for $\Delta\beta$ and $\Delta\rho$ are Benjamini–Hochberg FDR-adjusted within each contrast family (CIs unadjusted).

**Significance (two-sided bootstrap):** *** $p < .001$, ** $p < .01$, * $p < .05$, n.s. = not significant.

**“Single”** = single evaluation (no memory of prior trials). Study 2 was run for the two focal models (kimi-k2.5 and gpt-5-nano) across all three elicitation conditions.

## Supplementary Information

This Supplement reports the full per-model and per-condition parameter tables, the per-participant model-selection distributions, the choice-model goodness-of-fit and model-comparison tables, the loss-aversion and curvature-symmetry contrasts, the quality-control battery and base-model details, and the verbatim prompt texts. Unless otherwise stated, all confidence intervals are 95% percentile intervals from a cluster bootstrap that resamples whole simulated participants (n_boot = 2000), and significance is coded *** $p < .001$, ** $p < .01$, * $p < .05$, n.s. = not significant. P-values for the feeling-asymmetry contrasts are Benjamini–Hochberg FDR-adjusted within each contrast family; significance codes reflect the adjusted values (CIs unadjusted). Conditions are labelled single evaluation (`NO_MEMORY`), memory, and joint evaluation (`JOINT`); the two focal models (gpt-5-nano, kimi-k2.5) were run in all three conditions of both studies, and the remaining post-trained models in Study 1 single evaluation only.

### S1. Affect-rating value functions: full parameter tables

For every model × condition dataset we fit a single *pooled* fully asymmetric power-law value function to the ratings of all participants, $v(x) = \beta_{gain} \cdot x^{\rho_{gain}}$ for gains and $v(x) = -\beta_{loss} \cdot |x|^{\rho_{loss}}$ for losses, and obtained 95% confidence intervals on each parameter and on the gain − loss contrasts ($\Delta\beta$ for scale, $\Delta\rho$ for curvature) by the cluster bootstrap. $\beta$ indexes the scale of the rating function (its value at one unit, £1 or 1%) and $\rho$ its curvature ($\rho \ll 1$ = strong diminishing sensitivity; $\rho \approx 1$ = near-linear). A positive $\Delta\beta$ means gains are weighted more heavily than equivalent losses — the "silicon gain amplification" pattern. These per-arm estimates and contrasts are the values reported in the main text.

**Table S1.** Gain–loss asymmetry of the feeling (rating) function in Study 1 (abstract gambling), by model and elicitation condition. A compact version showing the focal-model rows only appears as main-text Table 1. Each row is a separate dataset (N = 30 simulated participants; 40 distinct outcomes, rated twice each in single evaluation for 80 rating trials and once each in the memory and joint conditions for 40). Estimates are from the pooled fully asymmetric fit; $\Delta\beta$ and $\Delta\rho$ are the gain − loss contrasts from the same fit with 95% cluster-bootstrap CIs. Positive $\Delta\beta$ = gains weighted more than losses. Significance is two-sided. $\rho_{loss}$ for gpt-4o-mini lies at the lower bound of the fit (0.01; §S9).

| Model | Condition | N | $\beta_{gain}$ | $\beta_{loss}$ | $\Delta\beta$ (scale) [95% CI] | $\rho_{gain}$ | $\rho_{loss}$ | $\Delta\rho$ (curvature) [95% CI] |
|---|---|---|---|---|---|---|---|---|
| claude-sonnet-4.5 | single | 30 | 25.0 | 17.2 | +7.83 [+7.78, +7.88] *** | 0.03 | 0.13 | −0.103 [−0.110, −0.097] *** |
| gemini-3-flash | single | 30 | 37.7 | 13.4 | +24.31 [+23.63, | 0.15 | 0.35 | −0.197 [−0.213, −0.181] *** |

| Model | Condition | N | β_gain | β_loss | Δβ (scale) [95% CI] | ρ_gain | ρ_loss | Δρ (curvature) [95% CI] |
|---|---|---|---|---|---|---|---|---|
| | | | | | +24.90] *** | | | |
| grok-4.3 | single | 30 | 27.8 | 21.3 | +6.48 [+6.06, +6.93] *** | 0.11 | 0.19 | −0.079 [−0.087, −0.069] *** |
| llama-3.1-70b | single | 30 | 22.0 | 23.3 | −1.32 [−1.87, −0.73] *** | 0.24 | 0.15 | +0.089 [+0.076, +0.101] *** |
| llama-4-maverick | single | 30 | 24.7 | 8.0 | +16.72 [+16.53, +16.92] *** | 0.22 | 0.37 | −0.145 [−0.155, −0.135] *** |
| gpt-4o-mini | single | 30 | 47.9 | 29.8 | +18.07 [+17.66, +18.46] *** | 0.03 | 0.01 | +0.018 [+0.014, +0.023] *** |
| gpt-5.2 | single | 30 | 14.0 | 12.7 | +1.24 [+0.98, +1.53] *** | 0.30 | 0.27 | +0.026 [+0.011, +0.040] *** |
| kimi-k2.5 | single | 30 | 18.4 | 15.3 | +3.09 [+2.81, +3.37] *** | 0.20 | 0.17 | +0.028 [+0.018, +0.038] *** |
| kimi-k2.5 | memory | 30 | 11.0 | 12.5 | −1.44 [−3.78, +0.56] n.s. | 0.56 | 0.53 | +0.036 [−0.043, +0.121] n.s. |
| kimi-k2.5 | joint | 30 | 4.6 | 5.9 | −1.37 [−2.27, −0.66] *** | 0.96 | 0.87 | +0.087 [+0.042, +0.133] *** |
| gpt-5-nano | single | 30 | 26.7 | 23.0 | +3.70 [+3.06, +4.31] *** | 0.16 | 0.17 | −0.018 [−0.034, −0.002] * |
| gpt-5-nano | memory | 30 | 20.8 | 16.2 | +4.63 [+1.10, +8.79] ** | 0.34 | 0.45 | −0.111 [−0.196, −0.029] ** |
| gpt-5-nano | joint | 30 | 1.6 | 1.6 | +0.01 [−0.00, +0.04] n.s. | 0.98 | 0.98 | −0.004 [−0.011, +0.000] n.s. |

Study 2 estimates (focal models only, all three conditions) are reported in main-text Table 2 and are not duplicated here.

Two patterns in Table S1 and main-text Tables 1–2 underpin the main-text claims. First, in single evaluation, ρ is far below 1 for every model (strong diminishing sensitivity) and Δβ is positive and significant for nearly all of them (gains weighted above losses); the one exception is llama-3.1-70b. Its smaller gain-side scale (Δβ = −1.32) is offset by a less compressed gain curve (Δρ = +0.089), so its gain and loss curves cross near £2, with losses rated more intensely below that amount and gains above it. Second, as elicitation moves from single evaluation to memory to joint, ρ rises towards 1 (linearisation) in both focal models and both studies, while the scale asymmetry Δβ shrinks, disappears or — in one Study 1 cell — reverses, becoming uniformly non-significant in Study 2. The asymmetry's response to comparison is model-dependent in Study 1 (gpt-5-nano remains gain-positive under memory and is abolished only under joint; kimi-k2.5 is non-significant under memory and reverses to a significant loss-side asymmetry under joint) but uniform in Study 2 (significant under single, non-significant under both memory and joint for both models). The robust, monotone effect of comparison is therefore linearisation of curvature, not a single fixed change in the asymmetry.

## S2. Per-participant value-function model selection

In addition to the pooled fits used for inference (§S1), we fit the full nested family of value functions to each participant individually and selected among them by BIC. The family comprises the linear model (single β, ρ ≡ 1), the linear-asymmetric model (separate β, ρ ≡ 1), the symmetric power-law model (shared β, ρ), the curvature-asymmetric ("Asymmetric") model (shared β, separate ρ), the scale-asymmetric model (separate β, shared ρ), and the fully asymmetric model (separate β and ρ). For the focal models and for the memory condition we additionally included a saturating model, $v(x) = \alpha \cdot \tanh(\beta x)$, which captures rating functions that rise approximately linearly and then floor/ceiling near the ends of the bounded scale. The counts below are from the fitting pass that includes the tanh model; the pooled β/ρ inference in §S1 uses the no-tanh baseline pass so that all datasets are compared on the same family. For gpt-5-nano in the Study 1 joint condition the counts are identical with and without the tanh candidate (Symmetric 25, Linear 5), so the main-text example is unaffected by the choice of pass.

The selection distributions track the linearisation story. Under single evaluation, scale-asymmetric forms (separate gain/loss scales) are the most common. Under memory, the saturating (tanh) and fully asymmetric forms take a substantial share, consistent with ratings that begin to compress against the scale ends as a running history accumulates. Under joint evaluation, the linear and symmetric forms become modal, consistent with near-linear, near-symmetric rating functions when all outcomes are compared in one pass.

**Table S2.** Best-fitting per-participant value-function model in Study 1, focal models, by condition (with-tanh fitting pass; counts out of 30 participants).

| Model | Condition | Best-fitting per-participant model (count) |
|---|---|---|
| kimi-k2.5 | single | ScaleAsymmetric 28; Asymmetric 2 |
| kimi-k2.5 | memory | FullAsymmetric 15; Tanh 7; Symmetric 4; ScaleAsymmetric 3; Linear 1 |
| kimi-k2.5 | joint | Linear 17; FullAsymmetric 10; Tanh 1; Symmetric 1; ScaleAsymmetric 1 |
| gpt-5-nano | single | ScaleAsymmetric 13; Symmetric 10; Tanh 6; Asymmetric 1 |
| gpt-5-nano | memory | FullAsymmetric 11; Tanh 9; Symmetric 4; Linear 3; ScaleAsymmetric 2; LinearAsymmetric 1 |
| gpt-5-nano | joint | Symmetric 25; Linear 5 |

**Table S3.** Best-fitting per-participant value-function model in Study 2, focal models, by condition (with-tanh fitting pass; counts out of 30 participants).

| Model | Condition | Best-fitting per-participant model (count) |
|---|---|---|
| kimi-k2.5 | single | FullAsymmetric 18; ScaleAsymmetric 12 |
| kimi-k2.5 | memory | Tanh 12; Symmetric 9; FullAsymmetric 4; Linear 3; Asymmetric 1; ScaleAsymmetric 1 |
| kimi-k2.5 | joint | Linear 28; Tanh 1; FullAsymmetric 1 |
| gpt-5-nano | single | ScaleAsymmetric 29; FullAsymmetric 1 |
| gpt-5-nano | memory | Tanh 10; Symmetric 7; Linear 6; FullAsymmetric 5; LinearAsymmetric 1; ScaleAsymmetric 1 |
| gpt-5-nano | joint | Linear 17; Symmetric 13 |

Across datasets the participant-level picture mirrors the pooled fits. Restricting to the nine Study 1 single-evaluation datasets — the basis of the main-text participant-level claim — $\beta_{gain}$ exceeded $\beta_{loss}$ in 210 of 270 per-participant fits (77.8%). Across all 13 Study 1 datasets (adding the focal models' memory and joint conditions, where the asymmetry has largely linearised away), the rate is 227 of 390 (58.2%), with the reverse pattern in 31 of 390 (7.9%) and a mean gain − loss scale difference of +6.11; curvature was close to symmetric (mean $\rho_{gain} - \rho_{loss} = -0.022$). In Study 2, $\beta_{gain}$ exceeded $\beta_{loss}$ in 60 of 60 single-evaluation fits (100%) and in 67 of 180 fits across all six datasets (37.2%; reverse pattern 10 of 180, 5.6%; mean scale difference +2.45; mean $\rho_{gain} - \rho_{loss} = -0.023$). Table S4 gives the per-dataset counts.

**Table S4.** Number of participants with β_gain > β_loss (per-participant power-law fits, no-tanh pass; 30 participants per dataset). Each participant is counted from their BIC-selected model; a selected model that shares a single β across gains and losses (β_gain = β_loss by construction) cannot satisfy the strict inequality, so shortfalls from 30 comprise both reverse-pattern and shared-β participants.

| Study | Model | Condition | β_gain > β_loss (of 30) |
|---|---|---|---|
| 1 | claude-sonnet-4.5 | single | 30 |
| 1 | gemini-3-flash | single | 30 |
| 1 | grok-4.3 | single | 30 |
| 1 | llama-3.1-70b | single | 0 |
| 1 | llama-4-maverick | single | 30 |
| 1 | gpt-4o-mini | single | 30 |
| 1 | gpt-5.2 | single | 15 |
| 1 | kimi-k2.5 | single | 28 |
| 1 | gpt-5-nano | single | 17 |
| 1 | kimi-k2.5 | memory | 6 |
| 1 | gpt-5-nano | memory | 11 |
| 1 | kimi-k2.5 | joint | 0 |
| 1 | gpt-5-nano | joint | 0 |
| 2 | kimi-k2.5 | single | 30 |
| 2 | gpt-5-nano | single | 30 |
| 2 | kimi-k2.5 | memory | 2 |
| 2 | gpt-5-nano | memory | 5 |
| 2 | kimi-k2.5 | joint | 0 |
| 2 | gpt-5-nano | joint | 0 |

## S3. Choice models: goodness of fit and model comparison

Choices were modelled as binary logits with the probability of choosing the gamble given by a softmax over option values, P(gamble) = $\sigma(\mu\cdot[V_{gamble} - V_{sure}])$. Five value models were compared: risk-neutral (expected value; k = 1), expected utility with constant relative risk aversion (CRRA; k = 2), prospect theory (single curvature α and loss-aversion weight λ; k = 3), full prospect theory (separate gain/loss curvatures; k = 4), and a feelings-informed model (option values derived from the participant's own fitted feeling function; k = 3 in Study 1 and 2 in Study 2, see §S5). For choice fits BIC = 2·NLL + k·ln(n). We report McFadden's pseudo-$R^2$ (relative to an intercept-only base-rate model) and choice-prediction accuracy (the fraction of trials on which the model's predicted majority choice matches the choice made) alongside BIC. A predicted probability of exactly 0.5 counts as predicting the gamble. The risk-neutral model predicts 0.5 whenever the gamble and the sure option have equal expected value, and the expected-utility model does so whenever a gamble's gain equals its loss. In each block the model with the lowest BIC is marked with an asterisk on its name.

**Table S5.** Choice-model goodness of fit, Study 1 (all 13 model × condition datasets).

| Model | Condition | Choice model | k | NLL | BIC | McFadden $R^2$ | Accuracy |
|---|---|---|---|---|---|---|---|
| claude-sonnet-4.5 | single | Risk Neutral | 1 | 1256.5 | 2520.9 | 0.329 | 0.742 |
| | | Expected Utility | 2 | 1071.2 | 2158.1 | 0.428 | 0.810 |
| | | Prospect Theory | 3 | 1061.7 | 2147.1 | 0.433 | 0.856 |
| | | **Full Prospect Theory*** | 4 | 1004.7 | 2041.0 | 0.463 | 0.865 |
| | | Feelings Informed | 3 | 1632.1 | 3287.8 | 0.128 | 0.701 |
| gemini-3-flash | single | Risk Neutral | 1 | 776.2 | 1560.2 | 0.583 | 0.812 |
| | | Expected Utility | 2 | 616.8 | 1249.5 | 0.669 | 0.894 |
| | | Prospect Theory | 3 | 555.2 | 1134.1 | 0.702 | 0.938 |
| | | **Full Prospect Theory*** | 4 | 531.9 | 1095.4 | 0.714 | 0.920 |
| | | Feelings Informed | 3 | 1618.6 | 3260.9 | 0.131 | 0.638 |
| gpt-4o-mini | single | Risk Neutral | 1 | 1759.3 | 3526.5 | -0.045 | 0.663 |
| | | Expected Utility | 2 | 1677.1 | 3370.0 | 0.004 | 0.719 |
| | | Prospect Theory | 3 | 1342.4 | 2708.6 | 0.203 | 0.746 |
| | | **Full Prospect Theory*** | 4 | 1324.5 | 2680.5 | 0.213 | 0.760 |
| | | Feelings Informed | 3 | 1379.1 | 2781.9 | 0.181 | 0.734 |
| gpt-5-nano | single | Risk Neutral | 1 | 435.8 | 879.4 | 0.767 | 0.817 |
| | | Expected Utility | 2 | 435.3 | 886.4 | 0.767 | 0.884 |
| | | Prospect Theory | 3 | 423.9 | 871.5 | 0.773 | 0.912 |

| Model | Condition | Choice model | k | NLL | BIC | McFadden $R^2$ | Accuracy |
|---|---|---|---|---|---|---|---|
| | | **Full Prospect Theory***| 4 | 352.3 | 736.2 | 0.812 | 0.951 |
| | | Feelings Informed | 3 | 1068.6 | 2161.0 | 0.428 | 0.822 |
| gpt-5-nano | memory | Risk Neutral | 1 | 340.8 | 688.8 | 0.635 | 0.828 |
| | | Expected Utility | 2 | 340.7 | 695.7 | 0.635 | 0.881 |
| | | Prospect Theory | 3 | 331.7 | 685.1 | 0.645 | 0.913 |
| | | **Full Prospect Theory***| 4 | 310.7 | 650.2 | 0.667 | 0.933 |
| | | Feelings Informed | 3 | 756.4 | 1534.4 | 0.190 | 0.706 |
| gpt-5-nano | joint | **Risk Neutral***| 1 | 258.3 | 523.8 | 0.719 | 0.881 |
| | | Expected Utility | 2 | 257.1 | 528.7 | 0.720 | 0.898 |
| | | Prospect Theory | 3 | 255.4 | 532.5 | 0.722 | 0.906 |
| | | Full Prospect Theory | 4 | 255.0 | 538.8 | 0.723 | 0.905 |
| | | Feelings Informed | 3 | 254.8 | 531.2 | 0.723 | 0.904 |
| gpt-5.2 | single | Risk Neutral | 1 | 686.8 | 1381.6 | 0.631 | 0.835 |
| | | Expected Utility | 2 | 551.6 | 1118.9 | 0.703 | 0.906 |
| | | Prospect Theory | 3 | 510.2 | 1044.1 | 0.726 | 0.952 |
| | | **Full Prospect Theory***| 4 | 498.3 | 1028.2 | 0.732 | 0.944 |
| | | Feelings Informed | 3 | 915.9 | 1855.5 | 0.507 | 0.847 |
| grok-4.3 | single | Risk Neutral | 1 | 941.3 | 1890.4 | 0.488 | 0.829 |
| | | Expected Utility | 2 | 659.8 | 1335.4 | 0.641 | 0.887 |
| | | **Prospect Theory***| 3 | 624.8 | 1273.3 | 0.660 | 0.931 |
| | | Full Prospect Theory | 4 | 624.4 | 1280.4 | 0.660 | 0.931 |

| Model | Condition | Choice model | k | NLL | BIC | McFadden $R^2$ | Accuracy |
|---|---|---|---|---|---|---|---|
| | | Feelings Informed | 3 | 1437.3 | 2898.3 | 0.218 | 0.740 |
| kimi-k2.5 | single | Risk Neutral | 1 | 1023.4 | 2054.7 | 0.439 | 0.810 |
| | | Expected Utility | 2 | 668.0 | 1351.8 | 0.634 | 0.885 |
| | | **Prospect Theory*** | 3 | 634.0 | 1291.6 | 0.652 | 0.931 |
| | | Full Prospect Theory | 4 | 633.4 | 1298.4 | 0.653 | 0.931 |
| | | Feelings Informed | 3 | 1231.2 | 2486.1 | 0.325 | 0.780 |
| kimi-k2.5 | memory | Risk Neutral | 1 | 825.5 | 1658.3 | 0.112 | 0.684 |
| | | Expected Utility | 2 | 705.3 | 1425.1 | 0.241 | 0.725 |
| | | Prospect Theory | 3 | 685.9 | 1393.3 | 0.262 | 0.759 |
| | | Full Prospect Theory | 4 | 684.6 | 1398.0 | 0.264 | 0.755 |
| | | **Feelings Informed*** | 3 | 481.6 | 984.8 | 0.482 | 0.863 |
| kimi-k2.5 | joint | Risk Neutral | 1 | 661.5 | 1330.2 | 0.293 | 0.781 |
| | | Expected Utility | 2 | 588.7 | 1191.8 | 0.371 | 0.808 |
| | | Prospect Theory | 3 | 500.0 | 1021.6 | 0.465 | 0.847 |
| | | **Full Prospect Theory*** | 4 | 495.0 | 1018.7 | 0.471 | 0.865 |
| | | Feelings Informed | 3 | 609.1 | 1239.8 | 0.349 | 0.825 |
| llama-3.1-70b | single | Risk Neutral | 1 | 1612.8 | 3233.5 | 0.135 | 0.641 |
| | | Expected Utility | 2 | 1505.9 | 3027.6 | 0.193 | 0.740 |
| | | Prospect Theory | 3 | 1494.5 | 3012.8 | 0.199 | 0.730 |
| | | **Full Prospect Theory*** | 4 | 1400.3 | 2832.2 | 0.249 | 0.733 |

| Model | Condition | Choice model | k | NLL | BIC | McFadden $R^2$ | Accuracy |
|---|---|---|---|---|---|---|---|
| | | Feelings Informed | 3 | 1460.8 | 2945.3 | 0.217 | 0.718 |
| llama-4-maverick | single | Risk Neutral | 1 | 1871.4 | 3750.6 | -0.241 | 0.412 |
| | | Expected Utility | 2 | 1769.5 | 3554.8 | -0.173 | 0.592 |
| | | Prospect Theory | 3 | 1768.4 | 3560.4 | -0.172 | 0.539 |
| | | Full Prospect Theory | 4 | 1696.3 | 3424.2 | -0.125 | 0.615 |
| | | **Feelings Informed*** | 3 | 1341.9 | 2707.5 | 0.110 | 0.749 |

**Table S6.** Choice-model goodness of fit, Study 2 (focal models × three conditions). Three cells (gpt-5-nano single and joint; kimi-k2.5 joint) produced identical choices on their seed-matched gamble sets, so their money-based rows coincide exactly; the feelings-informed rows, which use each cell's own ratings, differ. The duplicated rows are not transcription errors.

| Model | Condition | Choice model | k | NLL | BIC | McFadden $R^2$ | Accuracy |
|---|---|---|---|---|---|---|---|
| gpt-5-nano | single | Risk Neutral | 1 | 127.7 | 262.7 | 0.845 | 0.864 |
| | | Expected Utility | 2 | 127.7 | 269.8 | 0.845 | 0.864 |
| | | Prospect Theory | 3 | 11.0 | 43.6 | 0.987 | 1.000 |
| | | **Full Prospect Theory*** | 4 | 7.2 | 43.2 | 0.991 | 1.000 |
| | | Feelings Informed | 2 | 461.7 | 937.9 | 0.439 | 0.858 |
| gpt-5-nano | memory | Risk Neutral | 1 | 142.7 | 292.7 | 0.825 | 0.872 |
| | | Expected Utility | 2 | 136.2 | 286.7 | 0.833 | 0.872 |
| | | Prospect Theory | 3 | 60.8 | 143.3 | 0.925 | 0.990 |
| | | **Full Prospect Theory*** | 4 | 55.1 | 139.0 | 0.932 | 0.990 |
| | | Feelings Informed | 2 | 665.6 | 1345.7 | 0.182 | 0.925 |
| gpt-5-nano | joint | Risk Neutral | 1 | 127.7 | 262.7 | 0.845 | 0.864 |
| | | Expected Utility | 2 | 127.7 | 269.8 | 0.845 | 0.864 |
| | | Prospect Theory | 3 | 11.0 | 43.6 | 0.987 | 1.000 |

| Model | Condition | Choice model | k | NLL | BIC | McFadden $R^2$ | Accuracy |
|---|---|---|---|---|---|---|---|
| | | **Full Prospect Theory*** | 4 | 7.2 | 43.2 | 0.991 | 1.000 |
| | | Feelings Informed | 2 | 34.3 | 83.0 | 0.958 | 1.000 |
| kimi-k2.5 | single | Risk Neutral | 1 | 147.3 | 301.9 | 0.821 | 0.863 |
| | | Expected Utility | 2 | 147.0 | 308.4 | 0.822 | 0.863 |
| | | Prospect Theory | 3 | 30.6 | 82.8 | 0.963 | 0.999 |
| | | **Full Prospect Theory*** | 4 | 26.8 | 82.4 | 0.967 | 0.999 |
| | | Feelings Informed | 2 | 486.3 | 987.1 | 0.409 | 0.820 |
| kimi-k2.5 | memory | Risk Neutral | 1 | 595.6 | 1198.4 | 0.284 | 0.844 |
| | | Expected Utility | 2 | 481.3 | 977.1 | 0.422 | 0.844 |
| | | Prospect Theory | 3 | 473.8 | 969.2 | 0.431 | 0.891 |
| | | Full Prospect Theory | 4 | 470.6 | 970.1 | 0.434 | 0.891 |
| | | **Feelings Informed*** | 2 | 386.1 | 786.7 | 0.536 | 0.919 |
| kimi-k2.5 | joint | Risk Neutral | 1 | 127.7 | 262.7 | 0.845 | 0.864 |
| | | Expected Utility | 2 | 127.7 | 269.8 | 0.845 | 0.864 |
| | | Prospect Theory | 3 | 11.0 | 43.6 | 0.987 | 1.000 |
| | | **Full Prospect Theory*** | 4 | 7.2 | 43.2 | 0.991 | 1.000 |
| | | Feelings Informed | 2 | 56.0 | 126.4 | 0.932 | 0.996 |

A utility-based model — full prospect theory or, for two models, prospect theory — provides the best fit in the large majority of single-evaluation datasets, although the choices themselves largely follow expected value (Table S15). The prospect-theory models fit better because they also capture the departures from expected value and the preference for the sure option when expected values are equal. Two qualifications are visible in the tables and matter for interpretation. First, fit quality varies sharply across models: gpt-5-nano and gpt-5.2 are fit very well in single evaluation (McFadden $R^2 \approx 0.73$–$0.81$), whereas llama-4-maverick is fit poorly by every money-based model (negative McFadden $R^2$, near-chance accuracy), indicating noisy or weakly structured choices rather than a particular value shape. McFadden $R^2$ can be negative because the value models have no intercept, so they can fit worse than the base-rate model. Second, among the focal-model datasets the feelings-informed model attains the lowest BIC only in the memory condition of kimi-k2.5 (both studies) — the diagnostic case in which the model's own prior ratings are present in its context as it chooses (§S5). The single dataset among the single-evaluation models where the feelings-informed model nominally wins (llama-4-maverick) is not informative about affect entering choice, because in single evaluation the model has no access to its prior ratings; there the feelings-informed value simply acts as a flexible curve fitting an otherwise poorly-structured choice set (see §S7.2).

## S4. Loss aversion (λ) in choice

The loss-aversion parameter λ from (full) prospect theory measures loss aversion only in mixed gambles, in which a potential gain is set against a potential loss. Gain-only and loss-only trials cannot identify λ on their own. It does not enter gain-only trials, and in loss-only trials it multiplies the values of both options, so its effect cannot be separated from that of the inverse temperature μ. We therefore do not interpret λ from single-structure sub-fits. In the all-trial fits of Table S8, λ reflects the mixed gambles and also, through the shared μ, how consistently loss-only choices follow value relative to gain-only choices. A second identifiability caveat applies when choices are near-deterministic (prediction accuracy ≥ 0.98, i.e., fewer than 2% stochastic choices; the screen is applied identically in both studies): λ is then fixed by the deterministic choice rule rather than by graded trade-offs, so the point estimate and its interval are not interpretable even though they are numerically defined. In the near-deterministic Study 2 cells, for example, the gamble is never chosen when its gain equals its loss (184 of 1,350 trials in each of the three cells with identical choices), which any λ slightly above 1 reproduces. We flag such cells explicitly and parenthesise their λ values.

**Table S7.** Loss aversion in single-evaluation choice across all Study 1 models, from a **four-parameter full prospect-theory fit** (separate gain/loss curvatures α_gain, α_loss and a single λ), point estimates from the full-data fit (mixed gambles). For reference, human λ is typically ≈ 2; values here straddle 1 (near loss-neutrality). α_gain and α_loss are the gain/loss curvature parameters from the same fit.

| Model | α_gain | α_loss | λ (loss aversion) |
|---|---|---|---|
| claude-sonnet-4.5 | 0.967 | 0.911 | 1.192 |
| gemini-3-flash | 0.338 | 0.336 | 1.044 |
| grok-4.3 | 0.917 | 0.892 | 1.080 |
| llama-3.1-70b | 0.222 | 0.220 | 1.027 |
| llama-4-maverick | 0.441 | 0.306 | 1.585 |
| gpt-4o-mini | 0.272 | 0.233 | 0.880 |
| gpt-5.2 | 0.650 | 0.631 | 1.075 |
| kimi-k2.5 | 0.387 | 0.383 | 1.046 |
| gpt-5-nano | 0.512 | 0.503 | 1.028 |

**Table S8.** Loss aversion by condition, focal models, Study 1, from a **three-parameter prospect-theory fit** (single curvature α, single λ) to all choice trials (mixed, gain-only and loss-only), with cluster-bootstrap CIs over participants (n_boot = 2000; test against λ = 1, two-sided). Tables S7 and S8 differ in specification and in the trials used. Table S7 fits four

parameters to the mixed gambles only, the trials in which a gain is weighed against a loss. In the all-trial fit here, λ reflects the mixed gambles and also, through the shared inverse temperature, how consistently loss-only choices follow value relative to gain-only choices (§S4). The two tables therefore give different λ values for the same cell (e.g., 1.046 in Table S7 versus 1.070 here for kimi-k2.5 single evaluation). The main text quotes the three-parameter bootstrap values.

| Model | Condition | λ | 95% CI | p (H0: λ=1) | Sig. |
|---|---|---|---|---|---|
| gpt-5-nano | single | 1.005 | [1.002, 1.010] | < .001 | *** |
| gpt-5-nano | memory | 1.040 | [1.016, 1.074] | < .001 | *** |
| gpt-5-nano | joint | 1.009 | [1.001, 1.021] | 0.032 | * |
| kimi-k2.5 | single | 1.070 | [1.058, 1.083] | < .001 | *** |
| kimi-k2.5 | memory | 1.184 | [1.070, 1.308] | 0.001 | ** |
| kimi-k2.5 | joint | 1.277 | [1.185, 1.393] | < .001 | *** |

**Table S9.** Loss aversion by condition, focal models, Study 2, from the same **three-parameter prospect-theory fit** with cluster-bootstrap CIs as Table S8. Near-deterministic cells (choice accuracy ≥ 0.98) are not identifiable; their λ values are parenthesised and should not be interpreted or cited. The only interpretable Study 2 choice cell for loss aversion is kimi-k2.5 under memory.

| Model | Condition | Choice acc. | λ | 95% CI | p | Identifiable? |
|---|---|---|---|---|---|---|
| gpt-5-nano | single | 1.000 | (1.106) | [1.101, 1.115] | < .001 | no — near-deterministic |
| gpt-5-nano | memory | 0.990 | (1.028) | [1.020, 1.044] | < .001 | no — near-deterministic |
| gpt-5-nano | joint | 1.000 | (1.106) | [1.101, 1.115] | < .001 | no — near-deterministic |
| kimi-k2.5 | single | 0.999 | (1.106) | [1.038, 1.114] | < .001 | no — near-deterministic |
| kimi-k2.5 | memory | 0.891 | 1.007 | [1.002, 1.029] | 0.005 | yes |
| kimi-k2.5 | joint | 1.000 | (1.106) | [1.101, 1.115] | < .001 | no — near-deterministic |

Across single-evaluation models (Table S7), λ ranges from 0.88 (gpt-4o-mini) to 1.585 (llama-4-maverick), scattered around 1 — a level of loss aversion far below the human benchmark and, for one model (gpt-4o-mini), slightly below 1. In the focal-model breakdown (Table S8), λ is statistically distinguishable from 1 in all six cells but small in magnitude (for example gpt-5-nano λ ≈ 1.01–1.04). In Study 2 (Table S9), most cells are near-deterministic and uninterpretable; the single identifiable estimate, kimi-k2.5 under memory, is λ = 1.007 (95% CI [1.002, 1.029], p = .005) — reliably above 1 but negligibly so.

The choice data therefore do not reproduce the rating-side gain–loss asymmetry; choices are approximately loss-neutral, as noted in the main text.

## S5. Affect into choice: feelings-informed versus utility model comparison

To test whether a model's elicited affect entered its choices, the feelings-informed model values each option by passing its outcomes through the participant's own fitted feeling function, dividing the resulting values by the standard deviation of that participant's fitted feeling function over the 40 rated outcomes (without centring, so that a zero outcome keeps a subjective value of zero), and combining them with free non-negative weights on the gamble's gain side, loss side, and the sure option (inverse temperature fixed to 1, so the weights absorb scaling; $k = 3$ in Study 1; in Study 2 every sure option is a 0% change, so its weight is unidentifiable and $k = 2$). The feeling function used here is the participant's own per-participant fit from the **no-tanh** fitting pass, chosen so that the value basis is drawn from the same candidate family used for all pooled inference. These feeling parameters are estimated on the rating data alone and are held fixed when fitting choice; k therefore counts only the choice-stage weights and not the feeling-function parameters, so the feelings-informed model is not granted extra free parameters relative to the utility models it is compared against. We compared this model against the best utility-based model for each dataset by the difference in BIC, with a two-sided cluster-bootstrap p-value (see Methods, *Statistical analysis*), matching the two-sided contrasts in §S1, §S4 and §S6. The bootstrap holds each participant's feeling function at its full-data estimate, so these intervals are conditional on the fitted feeling functions and omit rating-stage estimation and model-selection uncertainty. In the tables below, ΔBIC = BIC(baseline utility model) − BIC(feelings-informed); a positive ΔBIC favours the feelings-informed model. The comparison is interpretable only in the memory condition, where the model's own prior ratings are in its context as it chooses; in single evaluation the model cannot consult prior ratings, so a feelings-informed fit there is a flexible curve, not evidence of affect-driven choice. As a further robustness check against the concern that the feelings-informed model wins only because its per-participant feeling functions absorb participant heterogeneity, we additionally compared it against a prospect-theory baseline fit separately to each participant ($k = 90$ in total), whose BIC charges each participant's three parameters against that participant's own trials (a blockwise penalty), since those parameters are informed by no other data. In single evaluation the per-participant baseline outperforms the feelings-informed model in all four focal cells (both models, both studies), consistent with no affect use in single-evaluation choice. Under memory it also outperforms the feelings-informed model in all four cells: narrowly for kimi-k2.5 (per-participant baseline vs feelings-informed BIC, Study 1 819.4 vs 984.8; Study 2 770.6 vs 786.7) and by a wide margin for gpt-5-nano (Study 1 754.5 vs 1534.4; Study 2 406.3 vs 1345.7). The pooled kimi-k2.5 advantage therefore does not rule out between-participant differences in monetary preferences as an explanation. In the joint condition this check is uninformative: the models behave as near-perfect expected-value maximisers ($P(\text{gamble} \mid EV > 0) \approx 0.99$–$1.00$ and $P(\text{gamble} \mid EV < 0) \approx 0.00$ for gpt-5-nano in Study 1 and for both models in Study 2), so both model families fit near-perfectly, the likelihood saturates, and the comparison is driven by the parameter penalties rather than by fit; in the one graded joint cell (kimi-k2.5, Study 1) the per-participant baseline wins outright. For the joint condition we therefore rely on the pooled comparison (Tables S10 and S11). The near-perfect expected-value thresholding under joint evaluation is itself notable: once ratings linearise, choice reduces to pure calculation.

**Table S10.** Affect-into-choice model comparison, focal models, Study 1. “Baseline” is the best utility model for that cell (FPT = full prospect theory, PT = prospect theory, RN = risk-neutral).

| Model | Condition | Baseline vs feelings-informed | ΔBIC | 95% CI | p (two-sided) |
|---|---|---|---|---|---|
| gpt-5-nano | single | FPT vs FI | -1424.7 | [-1558.9, -1262.7] | < .001 |
| gpt-5-nano | memory | FPT vs FI | -884.3 | [-1065.6, -679.1] | < .001 |
| gpt-5-nano | joint | RN vs FI | -7.4 | [-26.3, +14.0] | 0.475 |
| kimi-k2.5 | single | PT vs FI | -1194.5 | [-1395.2, -1007.6] | < .001 |
| kimi-k2.5 | memory | PT vs FI | +408.5 | [+257.7, +562.8] | < .001 |
| kimi-k2.5 | joint | FPT vs FI | -221.1 | [-365.6, -95.1] | 0.002 |

**Table S11.** Affect-into-choice model comparison, focal models, Study 2.

| Model | Condition | Baseline vs feelings-informed | ΔBIC | 95% CI | p (two-sided) |
|---|---|---|---|---|---|
| gpt-5-nano | single | FPT vs FI | -894.7 | [-1092.2, -607.1] | < .001 |
| gpt-5-nano | memory | FPT vs FI | -1206.7 | [-1445.8, -555.1] | < .001 |
| gpt-5-nano | joint | FPT vs FI | -39.8 | [-48.5, -32.0] | < .001 |
| kimi-k2.5 | single | FPT vs FI | -904.7 | [-987.4, -795.9] | < .001 |
| kimi-k2.5 | memory | PT vs FI | +182.5 | [-63.3, +390.5] | 0.163 |
| kimi-k2.5 | joint | FPT vs FI | -83.2 | [-141.8, -35.0] | < .001 |

The informative cells are the memory conditions. In Study 1, kimi-k2.5 under memory is the one cell in which the feelings-informed model decisively beats the best pooled utility model (ΔBIC = +408.5, 95% CI [+257.7, +562.8], p < .001), but not participant-specific prospect-theory fits (above), so between-participant differences in monetary preferences remain an alternative account. gpt-5-nano under memory shows the opposite — the utility model remains far ahead (ΔBIC = −884.3). In Study 2, the same kimi-memory comparison points the same way but weakly and non-significantly (ΔBIC = +182.5, 95% CI [−63.3, +390.5], $p = .163$). In all single-evaluation cells the utility model wins by a wide margin, as expected given that prior ratings are unavailable there. Taken together, these comparisons provide no robust evidence that elicited affect improves the account of choice beyond monetary utility.

## S6. Curvature symmetry (the α_gain − α_loss contrast)

The *reflection effect* is the approximately symmetric S-shape of the value function — concave over gains and convex over losses, i.e. α_gain and α_loss each below 1. The contrast reported below, α_gain − α_loss from full prospect theory, measures only the *departure from symmetric reflection*; it is not itself a reflection effect, and we find it to be small and, where significant, negative. We report it for completeness. The same near-determinism caveat as in §S4 applies: in cells with choice accuracy ≥ 0.98 the contrast is not identifiable and is parenthesised.

**Table S12.** Curvature-symmetry contrast α_gain − α_loss by condition, focal models, Study 1 (two-sided test against zero).

| Model | Condition | α_gain − α_loss | 95% CI | p | Sig. |
|---|---|---|---|---|---|
| gpt-5-nano | single | -0.019 | [-0.030, -0.009] | < .001 | *** |
| gpt-5-nano | memory | -0.077 | [-0.130, -0.035] | < .001 | *** |
| gpt-5-nano | joint | -0.007 | [-0.029, +0.010] | 0.432 | n.s. |
| kimi-k2.5 | single | -0.008 | [-0.021, +0.006] | 0.256 | n.s. |
| kimi-k2.5 | memory | +0.041 | [-0.044, +0.127] | 0.350 | n.s. |
| kimi-k2.5 | joint | -0.052 | [-0.095, -0.011] | 0.014 | * |

**Table S13.** Curvature-symmetry contrast α_gain − α_loss by condition, focal models, Study 2. Near-deterministic cells are not interpretable (parenthesised).

| Model | Condition | Choice acc. | α_gain − α_loss | 95% CI | p | Interpretable? |
|---|---|---|---|---|---|---|
| gpt-5-nano | single | 1.000 | (+0.040) | [+0.037, +0.044] | < .001 | no — near-deterministic |
| gpt-5-nano | memory | 0.990 | (+0.011) | [+0.006, +0.034] | < .001 | no — near-deterministic |
| gpt-5-nano | joint | 1.000 | (+0.040) | [+0.037, +0.044] | < .001 | no — near-deterministic |
| kimi-k2.5 | single | 0.999 | (+0.040) | [+0.037, +0.044] | < .001 | no — near-deterministic |
| kimi-k2.5 | memory | 0.891 | +0.003 | [-0.001, +0.021] | 0.176 | yes |
| kimi-k2.5 | joint | 1.000 | (+0.040) | [+0.037, +0.044] | < .001 | no — near-deterministic |

The contrast is at most 0.08 in magnitude (gpt-5-nano memory, −0.077). Where it is significant it is negative (gpt-5-nano single evaluation and memory, and kimi-k2.5 joint, in Study 1), and the only positive estimates in interpretable cells (kimi-k2.5 memory, in both studies) are non-significant. Curvature is, to a good approximation, symmetric across gains and losses; the asymmetry that the paper emphasises is in scale (β), not in curvature.

## S7. Quality control, the base model, and robustness

### S7.1 Quality-control battery

Parsed data were screened with a fixed battery of participant-level checks, run as diagnostics: response/parse rate (retain if ≥ 0.8 of expected trials parsed), rating range (values outside [−50, +50] flagged at the trial level), valence consistency (mean gain rating not negative, mean loss rating not positive), response variance (no single rating value exceeding 90% of a participant's ratings), monotonicity (Spearman correlation of magnitude with rating not below −0.3), and a "parrot" check (flagging participants for whom more than 50% of ratings simply echoed the stated magnitude). Each participant-level check defines an exclusion criterion; the rating-range check flags individual trials instead. In practice no participant met any criterion (below); all trials were retained in the primary analyses, and flagged trials were examined in the sensitivity analysis reported at the end of this section. Separately from this battery, every response that yielded an extracted value was stored verbatim, and a re-extraction audit of the stored responses across both studies (~93,000 rating rows and ~96,000 sequential-condition choice rows, including superseded and test runs that are not deposited; joint-condition choice lists were not re-audited) identified and corrected systematic extraction errors confined to grok-4.3 Study 1 single-evaluation ratings (25 of 2,400 trials), plus one value in the archived llama-3.1-405b choice data used only in the precision analysis, from which two unparseable rows were also removed; after correction the audit reproduces every audited value exactly (see Methods). The raw-vs-cleaned comparison below uses the no-tanh fitting pass on both sides.

In practice the data were clean at the participant level: **no participant was excluded in any of the 13 Study 1 datasets or any of the 6 Study 2 datasets**. Trial-level flags were negligible in almost every dataset (0% in the large majority). The two exceptions both involve gpt-5-nano in the joint condition, with 275 of 1,200 trials (22.9%) flagged by the parrot check in Study 1 and 143 of 1,200 (11.9%) in Study 2. In this cell, 25 of 30 LLM-participants in Study 1 (13 of 30 in Study 2) rated every outcome at face value, giving the signed amount rounded to a whole number (e.g. +12 for WIN £12 and −12 for LOSE £12). The other participants used a steeper linear scale. Because the parrot check flags a rating equal to the unsigned magnitude, it catches these participants' whole-number gains (11 of their 40 ratings each) but not the matching losses. Both response patterns are symmetric and essentially linear, so the flags mark legitimate ratings rather than echoed text. Excluding them (sensitivity analysis below) removes only the gain half of the pattern, which accounts for the small gain–loss difference after cleaning, and does not change the qualitative result for that cell (joint-condition ratings remain near-linear and near-symmetric). Following the extraction correction described in Methods, grok-4.3 single evaluation has no flagged trials: its earlier out-of-range and parrot flags were artefacts of the corrected extraction errors. Separately, a few sequential-condition datasets lost a handful of trials at the parsing stage rather than to the quality checks (for example kimi-k2.5 single evaluation retained 2,389 of 2,400 trials), with no material effect.

As a sensitivity analysis we re-fit every main-batch model × condition dataset after removing all flagged trials and compared the across-participant mean scale and curvature estimates before and after. For the only materially affected cells the estimates are effectively unchanged. For gpt-5-nano joint (Study 1), ($\beta_{gain}$, $\beta_{loss}$) moved from (1.590, 1.590) raw to (1.664, 1.521) cleaned and ($\rho_{gain}$, $\rho_{loss}$) from (0.969, 0.969) to (1.000, 1.000); in Study 2 the same cell moved from β (2.706, 2.706) to (2.745, 2.671) and ρ (0.984, 0.984) to (1.000, 1.000) — i.e. cleaning pushes the joint condition marginally *closer* to the exact linearity reported in the main text. The Study 2 gpt-5-nano memory cell had a single flagged (parrot) trial (1 of 1,200). All other main-batch datasets had no flagged trials and so were identical before and after (for the base-model dataset see §S7.3). Every qualitative conclusion — diminishing sensitivity in single evaluation, gain amplification, and the single < memory < joint linearisation — holds on both the raw and the quality-cleaned data.

### S7.2 The single-evaluation feelings-informed "win"

Among the single-evaluation models, the feelings-informed model attains the lowest choice BIC in exactly one dataset, llama-4-maverick (Table S5). This is not evidence that affect entered that model's choices. The feelings-informed model provides an additional flexible nonlinearity that fits an otherwise poorly-structured choice set (llama-4-maverick is the model that every money-based choice model fits poorly, with negative McFadden $R^2$). The interpretable affect-into-choice evidence comes only from the memory condition (§S5).

### S7.3 The base model (llama-3.1-405b)

The base (non-post-trained) model could not be run in the main batch because it was withdrawn by the provider before data collection. We therefore examined it in an earlier dataset (N = 60 simulated participants) collected under a slightly different Study 1 prompt that did not state the 50/50 win probability explicitly. In that dataset the base model showed both feeling signatures seen in the post-trained models: strong diminishing sensitivity ($\rho \approx 0.07$) and gain amplification (across-participant means of the per-participant fits: $\beta_{gain}$ = 36.9 vs $\beta_{loss}$ = 28.3; $\beta_{gain} > \beta_{loss}$ in 62% of participants), with gain amplification present more broadly across that dataset (mean $\beta_{gain} > \beta_{loss}$ in 9 of 10 single-evaluation datasets). Because the pattern appears in a model that has not undergone instruction tuning or alignment, and gain amplification is present under both the earlier and the corrected prompt wording, it is unlikely to be merely an artefact of post-training or of one prompt wording. We report these base-model values as preliminary descriptive evidence only: they come from a different prompt and sample, are confounded with both, and are not re-estimated, re-scaled, or standardised to the main batch. Of the 4,800 base-model rating trials, 4,626 (96.4%) parsed successfully. The quality battery flagged one of the 60 LLM-participants as meeting an exclusion criterion (inverted loss-side monotonicity, Spearman r = −0.36) and, at the trial level, one out-of-range rating and 191 echoed-magnitude ('parrot') trials among the other 59. Base-model results were essentially unchanged whether computed on the raw or the quality-cleaned data, and the gain–loss asymmetry persists under the stronger check of removing all flagged trials and refitting the per-participant models (mean predicted rating at ±£5: +41 vs −32 raw; +43 vs −33 flag-removed; Fig. 4). The earlier dataset reflects a model that is no longer available from the provider and is not regenerable.

### S7.4 Bounded-scale robustness

We did not run this check for the present submission; we record it here as a stated limitation and a natural confirmatory test. A reviewer may ask whether the curved, gain-skewed rating functions are an artefact of the bounded −50/+50 response scale: a finite scale mechanically compresses large magnitudes (mimicking diminishing sensitivity), and presenting many outcomes in one prompt may invite spreading ratings across the available range (mimicking linearisation). Relatedly, the joint condition changes the response format (a single pass returning a terse numbered list) as well as the comparative context; the memory condition, which holds the trial-by-trial format constant, provides the format-matched evidence for linearisation (see Methods). Several features of the present data argue against a pure scale artefact — a symmetric bounded scale does not predict the gain-vs-loss asymmetry, and the curvature replicates across two substantively different framings. A direct robustness check would re-run the rating task under (i) a wider or unbounded numeric scale and (ii) an explicitly anchored scale with fixed endpoints tied to reference outcomes, testing whether $\rho$ and $\Delta\beta$ are stable. We note, additionally, that the linearisation observed under joint presentation is the qualitative pattern predicted by range–frequency theory for judgements made over a presented set (Parducci, 1965; main-text ref. 24), which reframes it as a documented human judgment phenomenon rather than a measurement defect.

## S8. Prompt texts

All prompts are reproduced verbatim. Placeholders such as `{amount}`, `{gain}`, `{loss}` and `{sure}` denote the per-trial magnitudes drawn from the 20-point grid (Methods); `£` amounts (Study 1) appear as `%` price moves in Study 2. Chat/instruct models received the question-style prompts; the base model received the completion-style prompts, whose user text ends mid-sentence so that the continuation supplies the rating or choice; the delivered string wrapped this text in a minimal role scaffold, “SYSTEM: {system prompt}”, “USER: {prompt}” and a final “ASSISTANT:”, separated by blank lines. The base model was run in Study 1 only (single evaluation); no Study 2 or joint-condition base-model prompts were used, and none are listed here. The joint-condition prompts present a whole task in one message and request a numbered list. Two spacing artefacts in the delivered prompts are noted for exactness: in the Study 1 choice prompt for chat/instruct models, the final two sentences were concatenated without a separating space (shown as delivered below), and in the Study 1 rating prompt two spaces precede “How will you feel” (single-spaced below). The Study 2 prompts are unaffected.

**Memory-condition history injection (both studies):** In the memory condition, each trial’s prompt is preceded by the accumulated history inside the same user message: > HISTORY OF PREVIOUS TRIALS: > {history} > ======================================== > NEW TRIAL: > {trial prompt} — where {history} is the concatenation of “Question: {trial prompt}” / “Answer: {model response}” pairs from all preceding trials, separated by lines containing ---. In Study 1 both rating and choice trials append to the history; in Study 2 only rating trials do (see Methods).

Decoding settings were held fixed throughout: sampling temperature requested at 0.5, a 20-token response limit for the base (completion) model — sufficient for the short numeric rating or single letter its prompts elicit — and a 20,000-token limit for the chat/instruct models, which accommodates any reasoning trace the gateway returns before the parsed answer. Some reasoning models do not accept a temperature setting. The OpenAI reasoning models (gpt-5-nano, gpt-5.2) accept only their default temperature of 1.0, and kimi-k2.5’s developer documentation also lists its temperature as fixed. The gateway does not report the temperature applied, so the effective temperature of these three models may have differed from 0.5.

### System prompts

**Study 1 (all conditions):** > You are a participant in a decision-making study. You will be presented with choices involving financial outcomes. Please select the option that you would prefer.

**Study 2 (all conditions):** > You regularly trade stocks and shares on an app, but can't remember exactly which shares you own right now and which you have sold.

### Study 1 — sequential conditions (single evaluation and memory)

**Affect-rating task, chat/instruct models** (gain shown; the loss version replaces WIN with LOSE): > You are about to pick one of two shapes at random in a trial of an experiment. If you pick the correct shape (50% chance) you will WIN £{amount}, otherwise you'll get £0. How will you feel if you WIN £{amount}? Respond with a single number from -50 (Very Unhappy) to 50 (Very Happy). Do not add any other text.

**Affect-rating task, base model** (earlier wording, as delivered; gain shown; loss version replaces winning/WIN with losing/LOSE): > I am about to pick a shape at random. If I pick the correct shape I will WIN £{amount}, otherwise I'll get £0. On a scale from -50 (Very Unhappy) to +50 (Very Happy), the feeling rating I would give for winning £{amount} is

**Choice task, chat/instruct models:** > Option A: {option A text} > Option B: {option B text} > Which do you choose? Respond with just the option you want to pick. Do not add any other text.Wrap your final decision in tags, for example: <choice>Option A</choice> or <choice>Option B</choice>.

**Choice task, base model:** > Option A: {option A text} > Option B: {option B text} > Between these two options, I would choose Option

**Option templates (Study 1 choice).** The gamble appeared as Option A on 45 of 90 trials in the 90-trial design and on 22 of 45 in the 45-trial designs. > Mixed gamble: “50% WIN £{gain}, 50% LOSE £{loss}” versus “Guaranteed £0” > Gain-only: “50% WIN £{gain}, 50%

£0” versus “Guaranteed WIN £{sure}” > Loss-only: “50% LOSE £{loss}, 50% £0” versus “Guaranteed LOSE £{sure}”

### Study 1 — joint-evaluation condition

**Affect-rating task (single prompt).** Items are the interleaved WIN/LOSE outcomes, numbered: > You are taking part in an experiment made up of a series of independent trials. In each trial, you pick one of two shapes at random. If you pick the correct shape (50% chance), the outcome shown below happens; otherwise you get £0. > > For each trial below, rate how you would feel if the outcome shown happened. > > Use the scale: -50 (very unhappy) to +50 (very happy). > > Important: treat each trial as a separate, independent scenario. > > Respond with ONLY a numbered list of whole-number ratings in the same order. No explanation. Example format: > 1. 20 > 2. -15 > 3. 8 > > Trials: > 1. WIN £{amount} / 2. LOSE £{amount} / … (one line per outcome)

**Choice task (single prompt).** Each line presents one gamble as Option A versus Option B: > You will now make a series of independent choices. In each trial you choose between Option A and Option B. Gambles are decided by a fair 50/50 draw. > > For each trial below, respond with the letter of the option you would pick. > > Important: treat each trial as a separate, independent choice. > > Respond with ONLY a numbered list of letters (A or B) in the same order. No explanation. Example format: > 1. A > 2. B > 3. A > > Trials: > 1. Option A: {…} | Option B: {…} / … (one line per gamble)

### Study 2 — sequential conditions (single evaluation and memory)

**Affect-rating task, chat/instruct models** (direction is “up” for gains, “down” for losses): > While browsing your trading app, you notice a stock is {up/down} {amount}% today. You can't remember whether you own this stock — it's equally likely that you do or don't (50/50). How would you feel if you checked your portfolio and found out you do own it? Respond with a single number from -50 (Very Unhappy) to 50 (Very Happy). Do not add any other text.

**Choice task, chat/instruct models:** > A piece of news is being released tomorrow that will move a stock's price. If the news is good (50% chance), you estimate the stock will go up {gain}%. If the news is bad (50% chance), you estimate the stock will go down {loss}%. > > Option A: {Buy the stock / Do not buy (0% change)} > Option B: {the other option} > Which do you choose? Respond with just the option you want to pick. Do not add any other text. Wrap your final decision in tags, for example: <choice>Option A</choice> or <choice>Option B</choice>.

### Study 2 — joint-evaluation condition

**Affect-rating task (single prompt):** > You have been browsing your trading app and noticed a series of stock price changes today. For each stock, you can't remember whether you own it — it's equally likely that you do or don't (50/50). > > Below is a list of stock price changes. For each one, rate how you would feel if you checked your portfolio and found out you DO own that stock. > > Use the scale: -50 (very unhappy) to +50 (very happy). > > Important: treat each stock as a separate, independent scenario. > > Respond with ONLY a numbered list of whole-number ratings in the same order. No explanation. Example format: > 1. 20 > 2. -15 > 3. 8 > > Stocks: > 1. A stock is up {amount}% today. / 2. A stock is down {amount}% today. / …

**Choice task (single prompt):** > You are deciding whether to buy stocks ahead of news announcements being released tomorrow. For each stock, a piece of news will move its price — but you don't know which way. > > The alternative to buying is to not invest (0% change). > > For each stock below, respond with 1 if you would Buy, or 0 if you would Not Buy. > > Respond with ONLY a numbered list of 1s and 0s in the same order. No explanation. Example format: > 1. 1 > 2. 0 > 3. 1 > > Stocks: > 1. Good news (50% chance): up {gain}% | Bad news (50% chance): down {loss}% / …

The Study 2 joint choice format uses a fixed response mapping (1 = buy, 0 = do not buy): unlike the sequential choice tasks and the Study 1 joint condition, option positions are not counterbalanced, and no explicit instruction to treat trials as independent is included (the single-prompt list format was judged to make one unnecessary). The gamble_is_a column present in the deposited Study 2 joint-condition choice files did not influence these prompts.

### Note on corrected versus earlier prompt wording

The main analyses use a corrected version of the Study 1 prompts. An earlier version contained a formatting issue (missing separating spaces between adjacent sentences) and did not state the 50/50 win probability explicitly; the corrected prompts fix the system-prompt spacing and add the explicit "one of two shapes … (50% chance)" framing shown above, matching the choice task. All post-trained-model Study 1 data and all Study 2 data use the corrected prompts. The base-model dataset (§S7.3) was collected under the earlier wording and is reported separately and descriptively. The base model also received the earlier system prompt, delivered as "You are a participant in a decision-making study.You will be presented with choices involving financial outcomes.Please select the option that you would prefer" (no separating spaces or final full stop).

## S9. Optimisation and model-fitting details

All models — the affect-rating value functions, the choice models, and the feelings-informed model — were fit by maximum likelihood (least squares for the Gaussian rating models) using `scipy.optimize.minimize` with the L-BFGS-B algorithm under box constraints. To reduce sensitivity to local optima we used a multistart strategy: each fit was launched from a default starting point plus a set of additional starts drawn from a log-uniform distribution over the parameter bounds (uniform for the feelings-informed weights, whose lower bound is 0), and the solution with the lowest finite objective was retained. For the pooled rating fits and the choice fits we used on the order of 5–15 random starts per fit; each bootstrap resample was itself re-fit from 3 random starts, again keeping the best. Optimisation used fixed random seeds so that fits are reproducible, and each bootstrap iteration was seeded independently as a deterministic function of its index, so that resampling results are invariant to how the bootstrap is parallelised across processes.

Parameter bounds were as follows. *Feeling (rating) models:* scale parameters β on the power-law models were bounded to [0.01, 100]; the curvature exponents ρ to [0.01, 2]; the linear-model slope to [0.01, 20]; and, for the saturating tanh model $v(x) = \alpha \cdot \tanh(\beta x)$, α to [1, 60] and β to [0.0001, 5]. *Choice models:* the curvature α to [0.05, 3], the loss-aversion weight λ to [0.1, 10], and the softmax inverse temperature μ to [0.001, 100]. *Feelings-informed model:* the non-negative outcome weights (three in Study 1, two in Study 2) to [0, 200], with the inverse temperature fixed at $\mu = 1$ (so the weights absorb both relative importance and overall scaling). A resample on which no start returned a finite solution was skipped rather than substituted; the number of successful resamples was recorded and equalled the full n_boot = 2000 in every reported cell.

## S10. Sample-size and trial-count determination

The adequacy of the participant and trial counts of the main batch was evaluated by a bootstrap precision (stability) analysis carried out on the earlier Study 1 dataset (single-evaluation condition; N = 60 simulated participants per model; §S7.3). The logic is that of a non-parametric power analysis targeted at estimation precision rather than null-hypothesis power: for each candidate design size we resample the earlier dataset, re-estimate the design-level summaries of the key parameters, and record how their sampling variability shrinks as the design grows. Because the earlier dataset was collected under the older prompt wording (§S8), using it to evaluate the main-batch design assumes that the sampling-variance structure of the estimates carries over to the corrected prompts; this assumption was not tested directly.

Two design dimensions were varied independently. *Participants:* whole simulated participants were resampled with replacement at sizes N = 5, 10, 15, 20, 25, 30, 40, 50, 60 (100 iterations at each size). On each resample the feeling parameters (β_gain, β_loss, ρ_gain, ρ_loss) were summarised as the across-participant mean of the per-participant estimates (each participant's model class fixed at its full-data BIC selection, tanh excluded; the fitted estimates are resampled, not refit), and the choice parameters (α_gain, α_loss, λ) were re-estimated by refitting the pooled full prospect-theory model to the resampled participants' trials. *Trials:* holding participants fixed, the number of rating trials per participant was subsampled with replacement over 10–80 (sizes 10, 20, 40, 60, 80), refitting each participant's selected feeling-model class on the subsampled trials, and the number of choice trials over 18–78 (sizes 18, 39, 60, 78; stratified across the mixed / gain-only / loss-only structures, preserving the design's domain balance), refitting the pooled full prospect-theory model; the standard error of each design-level estimate was recorded as a function of trial count (50 iterations at each size). The adopted design (N = 30 participants; 80 rating and 90 choice trials in Study 1 single evaluation, 40 rating and 45 choice trials in the memory and joint conditions and throughout Study 2) sits on the flat portion of these stability curves, where adding participants or trials yields only marginal absolute reductions in parameter standard error, quantified below.

Figure S1 shows the resulting stability curves and Table S14 summarises them at the adopted design. Standard errors decline with design size at close to the $1/\sqrt{n}$ rate expected from sampling theory: at N = 30 participants, and at the reduced trial counts of the memory, joint and Study 2 designs, standard errors exceed their values at the maximum analysed design by a median of 44–46% across model × parameter cells (the median within each design dimension), close to the +41% implied by a 2:1 size ratio under $1/\sqrt{n}$ scaling (Table S14). The relevant question for design adequacy is therefore not whether the curves have reached an asymptote — under $1/\sqrt{n}$ scaling they do not — but whether the absolute precision at stake beyond the adopted design is material. It is not: re-estimating at the maximum design reduces each parameter's standard error by a median of 0.5% of the parameter's absolute value across model × parameter cells, by less than 5% in 94% of cells, and by at most 15.9%, with the largest values arising for the choice parameters (α, λ) of the models whose choices are least well fit by the money-based choice models (gpt-4o-mini, llama-4-maverick and the base model; cf. §S3, §S7.2). At N = 30, the bootstrap standard error of every feeling parameter is at most 3.4% (β) and 16.0% (ρ) of the parameter's value, small against the condition differences and gain–loss contrasts the study is designed to detect. The Study 1 single-evaluation trial counts (80 rating, 90 choice) equal or exceed the largest subsampled sizes, so their standard errors are bounded above by the full-trial values shown.

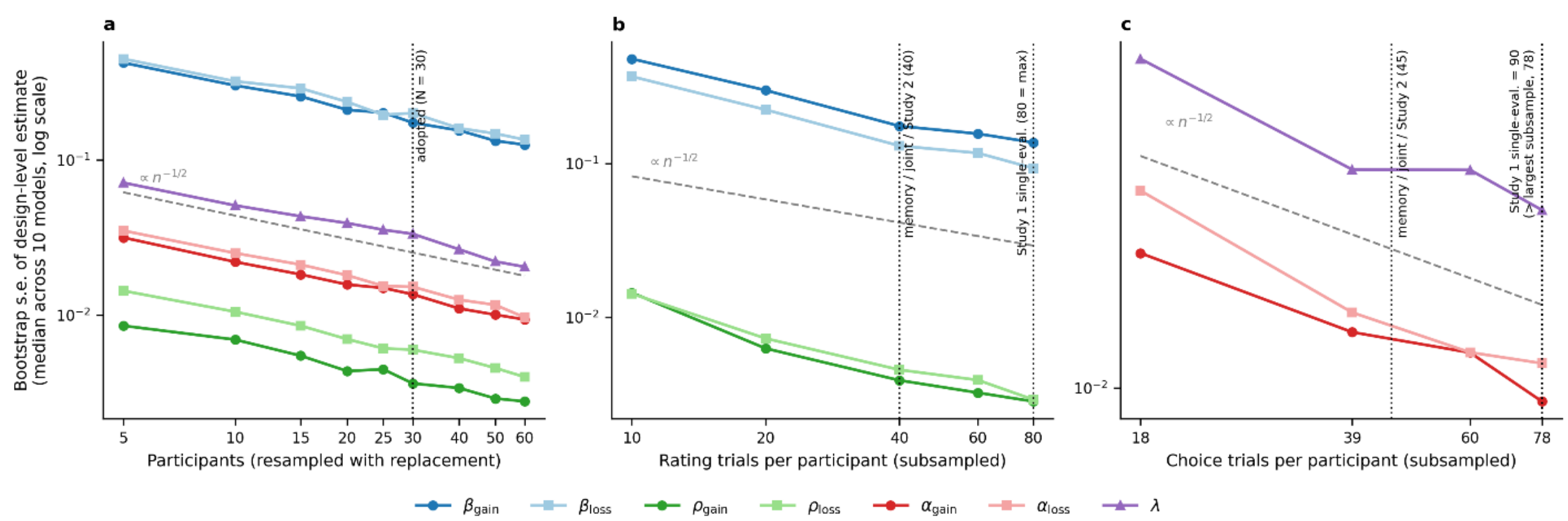

**Figure S1.** Parameter standard error as a function of design size, from the bootstrap precision analysis of the earlier Study 1 dataset. Each curve is the median across the 10 models of the bootstrap standard error of the design-level estimate — for the feeling parameters (β, ρ), the across-participant mean of per-participant fits; for the choice parameters (α, λ), the pooled full prospect-theory fit — on log–log axes; grey dashed lines show the $1/\sqrt{n}$ slope expected from sampling theory. (a) Participants resampled with replacement (5–60; 100 iterations per size). (b) Rating trials subsampled within participant (10–80; 50 iterations per size). (c) Choice trials subsampled within participant (18–78, stratified by trial structure; 50 iterations per size). Vertical dotted lines mark the adopted designs: N = 30 participants; 40 rating and 45 choice trials (memory, joint and Study 2); 80 rating trials (Study 1 single evaluation, equal to the maximum analysed). The Study 1 single-evaluation choice count (90) exceeds the largest subsampled size (78), so its standard error is bounded above by the 78-trial value.

**Table S14.** Design stability at the adopted sample and trial sizes (bootstrap precision analysis of the earlier Study 1 dataset; 10 models). The 10 models are those of the earlier dataset: the nine post-trained main-batch models, with grok routed as x-ai/grok-4.1-fast at the earlier collection date, plus the base model (llama-3.1-405b). "Adopted" is N = 30

(participant dimension) or the reduced per-participant trial counts of the memory, joint and Study 2 designs — 40 rating trials and, for choice, the 39-trial subsample, the largest analysed size below the adopted 45; "max" is the full earlier-dataset design (N = 60; 80 rating, 78 choice trials). s.e. columns give the median bootstrap standard error across models. "s.e. excess" is the percentage by which a model's standard error at the adopted size exceeds its value at the maximum, median (range) across models; values scatter around the +41% implied by $1/\sqrt{n}$ scaling for a 2:1 size ratio. "Precision cost" is the corresponding absolute increase in standard error expressed as a percentage of the parameter's value at the full design, median (maximum) across models.

| Design dimension (adopted vs max) | Parameter | s.e. at adopted (median) | s.e. at max (median) | s.e. excess at adopted, % — median (range) | Precision cost, % of parameter — median (max) |
|---|---|---|---|---|---|
| Participants (30 vs 60) | β_gain | 0.174 | 0.126 | +43 (+24 to +54) | 0.20 (0.7) |
| | β_loss | 0.201 | 0.135 | +42 (+25 to +60) | 0.32 (1.2) |
| | ρ_gain | 0.00362 | 0.00277 | +46 (+17 to +56) | 0.74 (3.7) |
| | ρ_loss | 0.00596 | 0.00398 | +43 (+28 to +76) | 1.35 (4.2) |
| | α_gain | 0.0136 | 0.00935 | +48 (+34 to +62) | 0.55 (11.5) |
| | α_loss | 0.0152 | 0.00965 | +55 (+11 to +78) | 0.63 (9.1) |
| | λ | 0.0334 | 0.0205 | +50 (+28 to +65) | 1.69 (12.5) |
| Rating trials (40 vs 80) | β_gain | 0.176 | 0.137 | +30 (−7 to +62) | 0.15 (0.4) |
| | β_loss | 0.131 | 0.0934 | +44 (+16 to +70) | 0.28 (0.6) |
| | ρ_gain | 0.00388 | 0.00283 | +40 (+10 to +67) | 0.81 (4.2) |
| | ρ_loss | 0.00455 | 0.00291 | +49 (+22 to +92) | 0.72 (4.7) |
| Choice trials (39 vs 78) | α_gain | 0.0132 | 0.00937 | +56 (+25 to +81) | 0.50 (7.8) |
| | α_loss | 0.0145 | 0.0113 | +38 (+11 to +82) | 0.36 (13.5) |
| | λ | 0.0294 | 0.0240 | +49 (+10 to +74) | 0.69 (15.9) |

## S11. Choice by expected value

**Table S15.** Gamble choice by expected value. Proportion of trials on which the gamble was chosen when it had a higher expected value than the sure option (EV > 0), a lower one (EV < 0) or the same (EV = 0), for each model × condition dataset, with the number of equal-expected-value trials. In Study 1 these comprise mixed gambles whose gain equals their loss and gain-only or loss-only gambles whose sure amount is half the risky amount; in Study 2 (mixed gambles only) they are gambles whose gain equals their loss. Values from section 4(c) of main_study.ipynb and study2_stock.ipynb.

| Study | Model | Condition | P(gamble), EV > 0 | P(gamble), EV < 0 | P(gamble), EV = 0 | EV = 0 trials |
|---|---|---|---|---|---|---|
| 1 | claude-sonnet-4.5 | single | 0.835 | 0.108 | 0.316 | 572 |
| 1 | gemini-3-flash | single | 0.969 | 0.056 | 0.264 | 572 |
| 1 | gpt-4o-mini | single | 0.829 | 0.536 | 0.566 | 572 |
| 1 | gpt-5-nano | single | 0.999 | 0.002 | 0.140 | 571 |
| 1 | gpt-5-nano | memory | 0.983 | 0.005 | 0.204 | 275 |
| 1 | gpt-5-nano | joint | 0.991 | 0.002 | 0.440 | 275 |
| 1 | gpt-5.2 | single | 0.992 | 0.026 | 0.278 | 572 |
| 1 | grok-4.3 | single | 0.972 | 0.086 | 0.381 | 572 |
| 1 | kimi-k2.5 | single | 0.943 | 0.100 | 0.381 | 567 |
| 1 | kimi-k2.5 | memory | 0.739 | 0.261 | 0.467 | 274 |
| 1 | kimi-k2.5 | joint | 0.819 | 0.017 | 0.389 | 275 |
| 1 | llama-3.1-70b | single | 0.730 | 0.212 | 0.227 | 572 |
| 1 | llama-4-maverick | single | 0.319 | 0.228 | 0.108 | 572 |
| 2 | gpt-5-nano | single | 1.000 | 0.000 | 0.000 | 184 |
| 2 | gpt-5-nano | memory | 0.999 | 0.000 | 0.065 | 184 |
| 2 | gpt-5-nano | joint | 1.000 | 0.000 | 0.000 | 184 |
| 2 | kimi-k2.5 | single | 0.999 | 0.000 | 0.000 | 184 |
| 2 | kimi-k2.5 | memory | 0.917 | 0.037 | 0.326 | 184 |
| 2 | kimi-k2.5 | joint | 1.000 | 0.000 | 0.000 | 184 |

# GUIDE-LLM: A reporting checklist for large language models in behavioural science

The GUIDE-LLM checklist provides a standardized framework for reporting studies that use large language models (LLMs) in the behavioral and social sciences. It aims to promote transparency, reproducibility, and ethical accountability across all stages of LLM-based research.

**How to complete the checklist:**
Fill in each item with brief, specific information about how LLMs were used in your study. Where an item does not apply, write “N/A” and, if helpful, note why. If multiple LLMs were used for different purposes, complete the relevant sections separately for each model. You may refer to sections or appendices in your manuscript rather than repeating text.

| Scope of LLM use | Answer |
| --- | --- |
| **Item A.1:** LLMs were used in this project for:<br>**Explanation:** Briefly describe how and for what purposes LLMs were used in the study. This may include one or multiple stages of the research workflow, depending on the project's design and aims. The following examples illustrate common use cases:<br>• **Research design** (e.g., hypothesis generation, literature search, or creating surveys/stimuli).<br>• **Data processing** (e.g., transcription, translation, data extraction, or data cleaning).<br>• **Analysis** (e.g., data labeling, summarization, pattern detection, statistical analysis, or code generation).<br>• **LLM as research object** (e.g., studying LLM behavior, benchmarking LLMs, or bias assessment of LLMs).<br>• **LLM as an evaluator** (e.g., evaluating model output, assessing the quality of human labeling/annotation)<br>• **Participant-facing settings** (e.g., LLM used as an intervention, studying human interactions with LLM chatbots).<br>• **Communication** (e.g., paper writing, editing, or reviewing).<br>Depending on the specific use case described here, different checklist items may later be relevant, and, in many cases, it may be necessary that later items in the checklist are reported separately for each use case. | (1) LLM as research object. Ten LLMs served as simulated participants ("LLM-participants") in an affect-rating task and a risky-choice task, in two studies and three elicitation conditions (Methods, Models and access; Simulated participants; Tasks and stimuli).<br>(2) Analysis (code generation) and communication. Claude (Anthropic) helped draft and edit the codebase and manuscript, helped the first author understand the AI-generated code it created, and was used to check the code against the manuscript. The authors reviewed, verified and take full responsibility for all content (Acknowledgements). |
| **Item A.2:** Degree of automation (human-in-the-loop vs. fully automated):<br>**Explanation:** Indicate how much human oversight was involved. Specify whether each output was reviewed, edited, or approved by a person, or whether outputs were used automatically without supervision. For participant-facing tasks, state whether humans checked outputs before showing them to participants or whether participants interacted with the LLM directly. Specify who provided oversight (e.g., student assistant, expert, PI). | (1) Data collection was automated. Seeded batch scripts sent independent API calls for each LLM-participant, and fixed parsers extracted ratings and choices with no human editing of model outputs. There were no human participants. The first author oversaw the code throughout. During collection the code printed the prompt text in the terminal for checking, and the stored data were checked against the expected outputs. After collection came an automated quality-control battery, a scripted re-extraction audit of the stored responses, and the first author's check of every corrected value (Methods, Response parsing and quality control; Supplementary §S7.1).<br>(2) All Claude output used in the project (code and text) was reviewed and approved by the authors. |

| Model/system details | Answer |
|---|---|
| **Item B.1:** Model name, including provider, model size, exact version/ID, date of access, and source link (if possible):<br><br>**Explanation:** Report the exact model names (including provider, version, and date accessed). Avoid generic labels like "ChatGPT" or "GPT-4"; instead, use detailed model names such as *"GPT-4o-mini-2024-12-17 (OpenAI)"* or *"Llama-3.1-8B (Meta; accessed via HuggingFace in May 2025)"*. For locally deployable models, please also enter a source link (e.g., the URL to the HuggingFace page). If multiple models were tested, it is encouraged to name them and briefly explain which one was used in the final study and why. When multiple models served different purposes, specify their respective roles, consistent with your response to Item A.1. | (1) Research object, accessed through OpenRouter (https://openrouter.ai/models) under these routed model strings. Post-trained models anthropic/claude-sonnet-4.5, google/gemini-3-flash-preview, x-ai/grok-4.3, meta-llama/llama-3.1-70b-instruct, meta-llama/llama-4-maverick, openai/gpt-4o-mini, openai/gpt-5.2, openai/gpt-5-nano and moonshotai/kimi-k2.5. Base (non-post-trained) model meta-llama/llama-3.1-405b. Model size is stated in the name only for llama-3.1-70b and llama-3.1-405b. Accessed 13 June 2026 (Study 1) and 16 June 2026 (Study 2). The base model was withdrawn before the main collection and comes from an earlier dataset (accessed 3 February 2026, in which the xAI model was x-ai/grok-4.1-fast). OpenRouter exposes no fixed snapshots, so model identity is the routed string plus access date (Methods, Models and access). gpt-5-nano and kimi-k2.5 were the focal models, run in every condition; the others were run in Study 1 single evaluation only.<br>(2) Claude (Anthropic), several versions, December 2025 to October 2026, for code and writing support. |
| **Item B.2:** Model access (e.g., API, web interface, local) and context mode (e.g., chat mode or separate calls):<br><br>**Explanation:** Note how you accessed the models (e.g., API, web interface, local installation) and whether you used LLMs in **chat mode** (ongoing conversation) or **stateless** mode (separate prompts). Mention the exact API name and version, since different access modes may influence responses (e.g., due to differences in model routing). | API through one gateway (OpenRouter), called with the OpenAI Python client 2.15.0. Chat/instruct models used the chat-completions endpoint and the base model the text-completions endpoint with completion-style prompts. Every call was stateless (one system and one user message). Context was varied by design. Single evaluation used one call per trial, memory placed the LLM-participant's own previous trials in each new prompt, and joint evaluation presented a whole task in one prompt (Methods, Elicitation conditions). No web interface or local model was used. |
| **Item B.3:** Relevant LLM configurations reported (as applicable), such as temperature, max tokens, seed, and number of runs:<br><br>**Explanation:** List any configuration settings that may affect outputs, such as:<br>● **temperature** (which controls randomness of the model's output)<br>● **Sampling parameters** such as **top_k, top_p, max tokens** (which limit the candidate token set or enforce length constraints)<br>● P**enalties** that discourage repetition (e.g., a frequency penalty to reduce the likelihood of tokens proportional to how often they have already appeared; a presence penalty reduces the likelihood of any token that has appeared at least once)<br>● **Stop sequences** (which halt generation when such a specified sequence is produced; e.g., ["\n\n", "END"]).<br>● **Number of completions or runs** (which is often used to capture variability in outputs across repeated generations)<br>● **Quantization level** (e.g., FP16, INT8, INT4) to change numerical precision beyond the default<br>● **Reasoning-related settings**, such as whether a specific structured reasoning was enabled, the specified reasoning effort level (e.g., low/medium/high or numerical settings that influence the depth of the reasoning), and any compute or inference budget constraints tied to the chosen reasoning mode | Temperature was requested at 0.5 for every model. The OpenAI reasoning models (gpt-5-nano, gpt-5.2) accept only their default of 1.0, kimi-k2.5's developer documentation lists its temperature as fixed, and the gateway does not report the value applied (Methods, Models and access). Maximum tokens 20,000 (chat/instruct) and 20 (base model). No top_p, top_k, penalties, stop sequences, seed or reasoning-effort settings were sent, so provider defaults applied; reasoning traces were stored where returned. One completion per call. Runs were 30 LLM-participants per model × condition (60 in the earlier dataset), each with a fixed seed that controls its stimuli and their order; the models' sampling was not seeded. Quantisation may vary by backend and is not reported by the gateway. |
| **Item B.4:** Customization:<br><br>**Explanation:** Check and describe any modifications or extended capabilities incorporated into your LLM setup beyond standard inference. This includes, but is not limited to:<br>● **Fine-tuning** (e.g., via LoRA; Low-Rank Adaptation) used to adapt a pretrained model to domain-specific data.<br>● **Retrieval-augmentation generation (RAG), where** the model retrieves relevant information from external sources (e.g., databases or document collections) during inference.<br>● **Automated prompt optimization** (e.g., DSPy) that treat prompts as trainable parameters.<br>● **Web search integration**, indicating whether the LLM was able to access and retrieve information from the Internet.<br>● **Agentic workflows**, including multi-step reasoning processes or delegated actions such as tool/function calling (e.g., via LangChain, AutoGPT, CrewAI).<br>● **Post-training refinements**, including alignment or optimization techniques used to adjust model behavior after pretraining (e.g., reinforcement learning from human feedback (RLHF), direct preference optimization (DPO)).<br>The goal is to specify any added customizations or provider-specific features that meaningfully shape system behavior in order to enable others to understand and accurately reproduce your setup. | ☒ Base model<br>☐ Fine-tuning<br>☐ RAG (retrieval-augmented)<br>☐ Automated prompt optimization<br>☐ Tool/function calling<br>☐ Web search<br>☐ Agentic workflows<br>☐ Other adaptations (e.g., safety mechanisms)<br><br>Description:<br>No customisation. All models were used as provided through the API, with no fine-tuning, retrieval, prompt optimisation, tools, web search or agentic workflows (Methods, Models and access). One of the ten models studied, llama-3.1-405b, was a base (non-post-trained) model. |
| **Item B.5:** Did the LLM session(s) include persistent memory across interactions?<br><br>**Explanation:** Indicate whether the LLM could "remember" previous conversations (i.e., had persistent memory). Unless such memory is disabled, there may also be spillover effects from other chat windows or prior conversations, which can influence outputs even when not intended. | ☐ Yes<br>☒ No<br>☐ N/A<br>No provider-side memory or session features were used, and every API call was independent. In the memory condition, each LLM-participant's own previous trials were written into its prompts (Methods, Elicitation conditions; format in Supplementary §S8). |

| **Prompts** | **Answer** |
|---|---|
| **Item C.1:** Exact prompt(s) reported:<br><br>**Explanation:** Whenever possible, include the **exact** text of prompts you used, including in-context examples or demonstrations provided to the LLM. Even small wording changes, formatting, or ordering of examples can substantially affect outputs. If full prompts cannot be shared (e.g., due to privacy or length), include a redacted or representative example or link to the full prompt in a repository (e.g., OSF, GitHub). If prompts were iteratively refined during the study, document the nature and extent of these refinements, including whether changes were pre-specified or made in response to interim outputs, to ensure transparency regarding potential optimization or selective reporting. | Yes. All prompts are reproduced verbatim in Supplementary §S8 (system prompts, chat and base-model rating and choice prompts, option templates, joint-condition list formats and the memory-condition history format), with two delivered spacing artefacts noted. The user prompt of every sequential-condition trial is also stored in the deposited data. One revision is documented. The earlier Study 1 wording, used for the base model and the precision analysis, lacked the explicit 50% probability and had missing spaces in the system prompt, which the main-batch prompts correct (Methods, Prompt wording and the base-model dataset). |
| **Item C.2:** System-wide instructions (if any):<br><br>**Explanation:** Note any *system-level* instructions that guide the model's general behavior (e.g., "You are a helpful assistant."). These are commonly not directly visible but can be accessed through the API. | Yes, reproduced verbatim in Methods and Supplementary §S8. Study 1 "You are a participant in a decision-making study. You will be presented with choices involving financial outcomes. Please select the option that you would prefer." Study 2 "You regularly trade stocks and shares on an app, but can't remember exactly which shares you own right now and which you have sold." |

| Data inputs & privacy | Answer |
| --- | --- |
| **Item D.1:** Handling of personal or sensitive data (if any) (e.g., consent for data processing):<br><br>**Explanation:** If any personal, sensitive, or identifiable data were processed, describe how they were handled in compliance with ethical standards and data protection laws. Researchers should indicate whether participants explicitly consented to their data being analyzed with an LLM, particularly when proprietary, cloud-based models are used. Such processing typically involves transferring data to a private company that may retain them indefinitely, which raises additional ethical and legal considerations. Beyond consent, describe how sensitive or identifiable data were handled (e.g., de-identification, anonymization, masking) and whether the LLM provider offers safeguards such as excluding inputs from training or storage. Clarify where data were stored or processed and how applicable legal/ethical requirements were met. If relevant, address cross-border transfers, as data may be stored in jurisdictions with different privacy laws (e.g., EU vs. US), with implications for compliance with GDPR, HIPAA, or other frameworks.<br><br>For context, some providers (e.g., OpenAI) may log or inspect prompts even when the data are not used for model training. For sensitive datasets, zero-retention configurations may be required (e.g., the MIMIC datasets can only be used with OpenAI models if a zero-retention checkpoint is enabled). | N/A. There were no human participants, and no personal or sensitive data were collected, processed or sent to model providers. Prompts contained only synthetic monetary outcomes from a fixed grid (Methods, Simulated participants). |

| Validation & interpretation | Answer |
| --- | --- |
| **Item E.1:** Human validation of LLM outputs:<br><br>**Explanation:** If relevant, describe whether and how human reviewers examined the model's outputs, and the degree of independence they had in doing so. Specify the reviewers' roles (e.g., domain experts, research assistants, subject-matter specialists) and relevant expertise, as well as how many reviewers participated and how their work was organized. Indicate whether outputs were independently annotated, double-checked by multiple reviewers, or merely approved or edited post-hoc by a lead author or investigator.<br><br>Clarify what dimensions of performance were examined. These may include known performance metrics from ML/AI such as accuracy or other metrics like citation correctness, hallucination detection, agreement or inter-rater reliability. State whether qualitative judgments, quantitative metrics, or both were used. If outcome assessment required subjective interpretation, describe assessor qualifications, instructions provided, and relevant demographics.<br><br>Describe the selection procedure for the reviewed outputs—whether all outputs were examined, a random sample was drawn, or specific cases (e.g., rare events or high-stakes responses) were oversampled to capture potential rare or critical errors. Further report how reviewers were trained or instructed, what criteria or rating scales they used, and how disagreements were resolved. For multi-reviewer settings, provide any inter-rater or inter-assessor reliability statistics (e.g., Cohen's κ or Krippendorff's α). Finally, note whether reviewer feedback was used purely for validation or also to refine prompts, retrain models, or adjust study procedures. | ☒ Yes<br>☐ No<br>☐ N/A<br><br>Two kinds of LLM output were reviewed, in both cases by the first author (PhD in experimental psychology, specialising in decision-making), working alone; there was no second, independent reviewer and no formal annotation.<br>(1) Research object. The model responses are the data under study, so no labelling was required. During data collection, responses were printed to the console as they were returned, and the first author monitored them during the runs and manually spot-checked a selection of rows in the saved CSV files to confirm that responses were well formed and correctly parsed. These checks were qualitative; systematic checking was automated.<br>(2) Code and writing support. All LLM-generated code and text were read and approved by the first author. Where needed, the first author also checked the outputs of the code against expected results, to confirm that it did what it was intended to do. |
| **Item E.2:** Describe any relevant post-processing (e.g., filtering in case of format mismatches, unit conversions, etc.):<br><br>**Explanation:** Describe any steps you took to clean or reformat LLM outputs (e.g., converting "positive/neutral/negative" to numeric codes, handling missing values, removing malformed entries). State how you handled inconsistent or unusable outputs and whether corrections were made with an automated script or manually. For example, when generating quantitative estimates (e.g., word counts, probabilities, or durations), the model may return values embedded in free text (e.g., "3.5 seconds") that require parsing and conversion into standardized numerical units. Post-processing steps should be described clearly, including how formatting errors, null responses, or inconsistent output structures were handled, whether automated scripts or manual corrections were used, and whether any data were excluded or reinterpreted as a result. Report how outputs were selected, extracted, or consolidated when multiple generations were produced. | All parsing was automated. Ratings were taken as the last number in a chat model's response and the first number in a base-model completion. Sequential choices used a fixed cascade (a <choice> tag, then the last "Option A/B" mention, then a trailing A/B), and the route for each trial was recorded. Joint-condition responses were parsed as numbered lists, and a list of the wrong length is discarded rather than guessed. Responses that yielded no value were not retained (33 rating and 39 choice rows of the main-batch design are absent). Extraction errors were corrected by script from the stored raw responses, and the originals are deposited. A sensitivity analysis removing all flagged trials left every qualitative conclusion unchanged (Methods; Supplementary §S7.1). |

| Reproducibility | Answer |
|---|---|
| **Item F.1:** Code/notebooks/scripts for LLM calls shared:<br><br>**Explanation:** Indicate whether you have shared materials such as code, prompts, logs, or transcripts. Make sure sensitive information (e.g., API keys, private data) is removed. For code, make sure to add a README file. | ☒ Yes<br>☐ No<br>☐ N/A<br><br>Link/DOI: https://github.com/joeamarks/Psychophysics_of_Affect_in_LLMs, archived at https://doi.org/10.5281/zenodo.23772352 (Code availability). The code includes the batch runners, agents, parsers and quality checks used for the API calls, the correction scripts and the analysis notebooks, with a README. The API key is read from a local .env file that is not shared. Data: https://doi.org/10.5281/zenodo.23556542 (Data availability). |

| Competing interests | Answer |
| --- | --- |
| **Item G.1:** Funding, support, or other relevant relationships (including in-kind access to compute or models, or professional affiliations):<br><br>**Explanation:** Disclose any current or past funding, support, or other relevant relationships with entities that have a financial interest in LLMs (this includes not just AI companies like OpenAI, Anthropic, but also tech companies developing or investing in AI, e.g., Google, Meta, Microsoft). This could include (but is not limited to): research funding from or collaborative research with a company with an interest in LLMs for this project or any other project within the past years; in-kind access to compute or models; current or former professional affiliations with a company with an interest in LLMs; personal investments (e.g., stocks) in companies with an interest in LLMs; familial relationship with an employee of a company with an interest in LLMs; etc. Disclose these relationships regardless of whether or not you believe they impacted the research. | ☒ Yes. Description: J.M. is employed by NatWest Group. The research was conducted in a personal capacity, and NatWest Group had no role in it (Competing interests). M.C.V. is supported by Swiss National Science Foundation grant PZ00P122329, which also paid for API access (approximately US$70 via OpenRouter; Acknowledgements). No funding, compute, model access or other support came from AI companies.<br>☐ No |

| Optional items | Answer |
|---|---|
| Justification for the specific LLM choice along the following dimensions:<br><br>**Explanation:** Briefly explain why you chose this specific model for your study and thereby explain why the selected model is suited to the research needs. Consider:<br>● **Performance:** Explain why this model's accuracy or size was appropriate for the research task, and, if relevant, consider compute cost, scalability, and real-time feasibility.<br>● **Transparency:** Was the model open-weight or closed-weight? Do we know what training data the model was exposed to or what the training objective was? If not, what are the implications for interpreting results? In terms of openness, different levels can be distinguished, in which only the model weights are publicly available ("open-weight") versus LLMs where the entire training process is available to allow for further customization and transparency of the LLM itself ("open-source").<br>● **Reproducibility:** Can the model be shared (so results can be reproduced in the future), or is access limited to proprietary/cloud services (so the access to the model may be altered, restricted, or discontinued)?<br>● **Ethical considerations:** Are there ethical reasons (including safety) for the model choice, such as running a model locally to avoid sharing sensitive data?<br>● **Other determinants** can be the cost or ease-of-use (e.g., due to accessibility of certain models via university infrastructures), if their benefits outweigh trade-offs in other dimensions. | ☒ Performance – Description: gpt-5-nano belonged to OpenAI's frontier model family at the time of the initial data collection and was efficient enough for the long memory-condition runs.<br><br>☐ Transparency – Description:<br><br>☐ Reproducibility – Description:<br><br>☐ Ethical considerations – Description:<br><br>☒ Others (e.g., cost, ease-of-use): – Description: kimi-k2.5 was at the top of OpenRouter's model list at the time (https://openrouter.ai/models) and seemed sufficiently different from gpt-5-nano to serve as the second focal model. The other seven post-trained models were chosen to represent the main developers' models at different capability levels. The base model tests whether the patterns precede post-training (Methods). |
| Discussion of the rationale for the prompt design:<br>**Explanation:** Explain how you designed your prompts. For example, indicate whether you used a structured format (e.g., explicit task description, definitions, step-by-step instructions, and output constraints), followed established prompt engineering guidelines or prior literature, adapted prompts from earlier studies, or relied on automated prompt optimization tools. Clarify whether the design was iterative (e.g., refined through pilot testing or error analysis), whether few-shot examples were included and how they were selected, and whether prompts were standardized across models to ensure comparability. | ☒ Yes. Description: The tasks adapt the design of Charpentier et al. (ref. 9) for LLMs, with a short participant-framing system prompt. Answer formats were constrained (a single number, a <choice> tag, or a numbered list) so that answers could be parsed automatically. Apart from answer-format examples, no examples were given: the sequential choice prompts show the tag format, and the joint-condition prompts show a three-line format example with illustrative answers (for ratings, 20, −15 and 8) that are not tied to any trial (Supplementary §S8). The same prompts were used for every chat model, and the base model received completion-style versions. One revision (explicit 50% probability and system-prompt spacing) is documented (Methods, Prompt wording and the base-model dataset).<br>☐ No<br>☐ N/A |
| Comparison against other methods/LLMs:<br>**Explanation:** If you compared different models, prompts, or methods, describe how you made the comparison and what criteria you used (e.g., accuracy). State whether you used the same inputs across methods and how you ensured fair comparison. If you considered but excluded alternative methods/LLMs, briefly explain why. If such comparisons are made, report the exact benchmarking procedure and the level of agreement between methods. If other LLMs or methods were considered but not used, briefly explain the rationale (e.g., cost, access, performance, or ethical considerations). | ☒ Yes. Description: Ten LLMs were compared on identical seeded stimuli, with the same estimators applied to every dataset. Estimates are set against published human benchmarks (curvature $\alpha \approx 0.88$ and loss aversion $\lambda \approx 2.25$, ref. 19; $\lambda \approx 1.96$, ref. 20) and against the human pattern in the same paradigm (ref. 9). Competing rating and choice models were compared by BIC, with cluster-bootstrap inference (Methods).<br>☐ No<br>☐ N/A |
| Training data leakage risks addressed:<br><br>**Explanation:** Note whether you considered the chance that your evaluation materials (e.g., test data, questionnaires) were already part of a model's training data. If relevant, describe steps taken to mitigate or detect leakage, such as novelty checks, re-phrasing or re-structuring items, or using hold-out materials not widely available. | ☒ Yes<br>☐ No<br>☐ N/A<br>The Charpentier et al. paradigm is published and could appear in training data. Leakage is nonetheless unlikely to explain the results, because the model ratings did not reproduce the published human data from this paradigm: Charpentier et al.'s participants rated gains and losses roughly equally (ref. 9), whereas the models rated gains more intensely than equivalent losses. Gain amplification inverts the canonical loss-aversion pattern that the human literature would supply, replicates in a new stock-trading framing (Study 2), was present in 9 of 10 models of the earlier dataset under the older wording, and also appears in a base model. Testing the link to training-corpus statistics is left to future work (Discussion). |
| Potential risk of bias or other systematic differences in LLM behavior that could affect the study's conclusions?<br><br>**Explanation:** Reflect on whether the LLM might perform differently across groups, languages, or contexts in ways that could affect your conclusions. If you checked for or mitigated such bias, describe how (e.g., subgroup analyses, balanced sampling, sensitivity checks). Where applicable, researchers may consult dedicated bias and fairness checklists for structured guidance on identifying and addressing these issues. | ☒ Yes. Description: Discussed in the limitations (Discussion) and Supplementary §S7. Gateway routing and quantisation can vary, and effective sampling temperatures were not reported and may differ between models. The bounded rating scale could shape curvature (§S7.4). The comparative conditions were run only for the two focal models.<br>☐ No |
| Conversation transcripts:<br><br>**Explanation:** For studies where participants or researchers interacted directly with an LLM (e.g., chat-based interventions, behavioral experiments, or coding tasks), provide anonymized transcripts or representative examples of the interactions. Redact any personally identifiable or sensitive information. | ☒ Yes, shared without sensitive information. Location: Deposited per-trial data (https://doi.org/10.5281/zenodo.23556542), with the stored user prompt, raw response and, in the sequential conditions, any reasoning trace for every trial that yielded a value. System prompts and joint-condition prompts are given in Supplementary §S8. The files contain no sensitive information.<br>☐ No. Reason for not sharing:<br>☐ N/A |
| Discuss relevant ethical implications of the research:<br><br>**Explanation:** Briefly discuss the broader ethical considerations of your study beyond procedural compliance (e.g., IRB approval). Address potential risks and safeguards, including:<br>● **Participant well-being:** foreseeable harms (emotional, reputational, financial, environmental), inclusion/exclusion criteria, effects on vulnerable groups, accessibility, use of deception and debriefing, autonomy and informed consent (including clear disclosure of AI involvement).<br>● **Fairness and equity:** whether outputs or outcomes may reflect or amplify bias or discrimination; note any subgroup analyses or mitigation strategies.<br>● **Safety:** broader risks (such as from agentic or autonomous behavior), or safeguards for harmful or adverse outputs (e.g., moderation layers, content filters, human-in-the-loop review, escalation protocols, red-teaming, dual use). | No human participants were involved. The Discussion considers what malleable, elicitation-dependent valuation means for LLM agents that advise on or make financial decisions. An expected-value-maximising agent, one that uses its own gain-skewed affective mapping, and one that adopts a user's preferences are each misaligned in a different way, and the choice between them is a normative question (Discussion). |
| Computational resources (e.g., API call counts, tokens, financial costs, or compute time):<br><br>**Explanation:** Report the computational resources required for the study. This may include API call counts, total tokens processed, financial costs, runtime, number and type of GPUs/CPUs used, and any other relevant compute metrics. | All models were accessed through the OpenRouter API, at a total cost of approximately US$70 for data collection in both studies (Acknowledgements). No local model hosting or GPUs were used. Analyses ran in Python 3.13.5 (NumPy 2.1.3, SciPy 1.15.3, pandas 2.2.3), with the cluster bootstraps run in parallel through joblib 1.4.2. The bootstraps take many hours (README). |

**How to cite**:
Feuerriegel, S Barrie, C., Crockett, M. J., Globig, L. K., McLoughlin, K. L., Mirea, D.-M., Spirling, A., Yang, D., ..., Rathje, S., & Ribeiro, M. H. (2026). *A reporting checklist for large language models in behavioural science*. In: *Nature Human Behaviour*.. DOI: 10.1038/s41562-026-02492-7